# An Introductory Study on Time Series Modeling and Forecasting

**Ratnadip Adhikari**

**R. K. Agrawal**

# ACKNOWLEDGEMENT

The timely and successful completion of the book could hardly be possible without the helps and supports from a lot of individuals. I will take this opportunity to thank all of them who helped me either directly or indirectly during this important work.

First of all I wish to express my sincere gratitude and due respect to my supervisor **Dr. R.K. Agrawal**, Associate Professor SC & SS, JNU. I am immensely grateful to him for his valuable guidance, continuous encouragements and positive supports which helped me a lot during the period of my work. I would like to appreciate him for always showing keen interest in my queries and providing important suggestions.

I also express whole hearted thanks to my friends and classmates for their care and moral supports. The moments, I enjoyed with them during my M.Tech course will always remain as a happy memory throughout my life.

I owe a lot to my mother for her constant love and support. She always encouraged me to have positive and independent thinking, which really matter in my life. I would like to thank her very much and share this moment of happiness with her.

Last but not the least I am also thankful to entire faculty and staff of SC & SS for their unselfish help, I got whenever needed during the course of my work.

RATNADIP ADHIKARI

# ABSTRACT

Time series modeling and forecasting has fundamental importance to various practical domains. Thus a lot of active research works is going on in this subject during several years. Many important models have been proposed in literature for improving the accuracy and effeciency of time series modeling and forecasting. The aim of this book is to present a concise description of some popular time series forecasting models used in practice, with their salient features. In this book, we have described three important classes of time series models, viz. the stochastic, neural networks and SVM based models, together with their inherent forecasting strengths and weaknesses. We have also discussed about the basic issues related to time series modeling, such as stationarity, parsimony, overfitting, etc. Our discussion about different time series models is supported by giving the experimental forecast results, performed on six real time series datasets. While fitting a model to a dataset, special care is taken to select the most parsimonious one. To evaluate forecast accuracy as well as to compare among different models fitted to a time series, we have used the five performance measures, viz. MSE, MAD, RMSE, MAPE and Theil's U-statistics. For each of the six datasets, we have shown the obtained forecast diagram which graphically depicts the closeness between the original and forecasted observations. To have authenticity as well as clarity in our discussion about time series modeling and forecasting, we have taken the help of various published research works from reputed journals and some standard books.

# CONTENTS

# List of Figures

# Introduction

Time series modeling is a dynamic research area which has attracted attentions of researchers community over last few decades. The main aim of time series modeling is to carefully collect and rigorously study the past observations of a time series to develop an appropriate model which describes the inherent structure of the series. This model is then used to generate future values for the series, i.e. to make forecasts. Time series forecasting thus can be termed as the act of predicting the future by understanding the past [31]. Due to the indispensable importance of time series forecasting in numerous practical fields such as business, economics, finance, science and engineering, etc. [7, 8, 10], proper care should be taken to fit an adequate model to the underlying time series. It is obvious that a successful time series forecasting depends on an appropriate model fitting. A lot of efforts have been done by researchers over many years for the development of efficient models to improve the forecasting accuracy. As a result, various important time series forecasting models have been evolved in literature.

One of the most popular and frequently used stochastic time series models is the *Autoregressive Integrated Moving Average (ARIMA)* [6, 8, 21, 23] model. The basic assumption made to implement this model is that the considered time series is linear and follows a particular known statistical distribution, such as the normal distribution. ARIMA model has subclasses of other models, such as the *Autoregressive (AR)* [6, 12, 23], *Moving Average (MA)* [6, 23] and *Autoregressive Moving Average (ARMA)* [6, 21, 23] models. For seasonal time series forecasting, Box and Jenkins [6] had proposed a quite successful variation of ARIMA model, viz. the *Seasonal ARIMA (SARIMA)* [3, 6, 23]. The popularity of the ARIMA model is mainly due to its flexibility to represent several varieties of time series with simplicity as well as the associated Box-Jenkins methodology [3, 6, 8, 23] for optimal model building process. But the severe limitation of these models is the pre-assumed linear form of the associated time series which becomes inadequate in many practical situations. To overcome this drawback, various non-linear stochastic models have been proposed in literature [7, 8, 28]; however from implementation point of view these are not so straight-forward and simple as the ARIMA models.

Recently, artificial neural networks (ANNs) have attracted increasing attentions in the domain of time series forecasting [8, 13, 20]. Although initially biologically inspired, but later on ANNs have been successfully applied in many different areas, especially for forecasting and classification purposes [13, 20]. The excellent feature of ANNs, when applied to time series forecasting problems is their inherent capability of non-linear modeling, without any presumption about the statistical distribution followed by the observations. The appropriate model is adaptively formed based on the given data. Due to this reason, ANNs are data-driven and self-adaptive by nature [5, 8, 20]. During the past few years a substantial amount of research works have been carried out towards the application of neural networks for time series modeling and forecasting. A state-of-the-art discussion about the recent works in neural networks for tine series forecasting has been presented by Zhang et al. in 1998 [5]. There are various ANN forecasting models in literature. The most common and popular among them are the multi-layer perceptrons (MLPs), which are characterized by a single hidden layer *Feed Forward Network (FNN)* [5,8]. Another widely used variation of FNN is the *Time Lagged Neural Network (TLNN)* [11, 13]. In 2008, C. Hamzacebi [3] had presented a new ANN model, viz. the *Seasonal Artificial Neural Network (SANN)* model for seasonal time series forecasting. His proposed model is surprisingly simple and also has been experimentally verified to be quite successful and efficient in forecasting seasonal time series. Offcourse, there are many other existing neural network structures in literature due to the continuous ongoing research works in this field. However, in the present book we shall mainly concentrate on the above mentioned ANN forecasting models.

A major breakthrough in the area of time series forecasting occurred with the development of Vapnik's support vector machine (SVM) concept [18, 24, 30, 31]. Vapnik and his co-workers designed SVM at the AT & T Bell laboratories in 1995 [24, 29, 33]. The initial aim of SVM was to solve pattern classification problems but afterwards they have been widely applied in many other fields such as function estimation, regression, signal processing and time series prediction problems [24, 31, 34]. The remarkable characteristic of SVM is that it is not only destined for good classification but also intended for a better generalization of the training data. For this reason the SVM methodology has become one of the well-known techniques, especially for time series forecasting problems in recent years. The objective of SVM is to use the *structural risk minimization (SRM)* [24, 29, 30] principle to find a decision rule with good generalization capacity. In SVM, the solution to a particular problem only depends upon a subset of the training data points, which are termed as the *support vectors* [24, 29, 33]. Another important feature of SVM is that here the training is equivalent to solving a linearly constrained quadratic optimization problem. So the solution obtained by applying SVM method is always unique and globally optimal, unlike the other traditional stochastic or neural network methods [24]. Perhaps the most amazing property of SVM is that the quality and complexity of the solution can be independently controlled, irrespective of the dimension of the input space [19, 29]. Usually in SVM applications, the input points are mapped to a high dimensional feature space, with the help of some special functions, known as *support vector kernels* [18, 29, 34], which often yields good generalization even in high dimensions. During the past few years numerous SVM forecasting models have been developed by researchers. In this book, we shall present an overview of the important fundamental concepts of SVM and then discuss about the *Least-square SVM (LS-SVM)* [19] and *Dynamic Least-square SVM (LS-SVM)* [34] which are two popular SVM models for time series forecasting.

The objective of this book is to present a comprehensive discussion about the three widely popular approaches for time series forecasting, viz. the stochastic, neural networks and SVM approaches. This book contains seven chapters, which are organized as follows: Chapter 2 gives an introduction to the basic concepts of time series modeling, together with some associated ideas such as stationarity, parsimony, etc. Chapter 3 is designed to discuss about the various stochastic time series models. These include the Box-Jenkins or ARIMA models, the generalized ARFIMA models and the SARIMA model for linear time series forecasting as well as some non-linear models such as ARCH, NMA, etc. In Chapter 4 we have described the application of neural networks in time series forecasting, together with two recently developed models, viz. TLNN [11, 13] and SANN [3]. Chapter 5 presents a discussion about the SVM concepts and its usefulness in time series forecasting problems. In this chapter we have also briefly discussed about two newly proposed models, viz. LS-SVM [19] and DLS-SVM [34] which have gained immense popularities in time series forecasting applications. In Chapter 6, we have introduced about ten important forecast performance measures, often used in literature, together with their salient features. Chapter 7 presents our experimental forecasting results in terms of five performance measures, obtained on six real time series datasets, together with the associated forecast diagrams. After completion of these seven chapters, we have given a brief conclusion of our work as well as the prospective future aim in this field.

# Basic Concepts of Time Series Modeling

### 2.1 Definition of A Time Series

A time series is a sequential set of data points, measured typically over successive times. It is mathematically defined as a set of vectors $x(t), t = 0,1,2,...$ where $t$ represents the time elapsed [21, 23, 31]. The variable $x(t)$ is treated as a random variable. The measurements taken during an event in a time series are arranged in a proper chronological order.

A time series containing records of a single variable is termed as univariate. But if records of more than one variable are considered, it is termed as multivariate. A time series can be continuous or discrete. In a continuous time series observations are measured at every instance of time, whereas a discrete time series contains observations measured at discrete points of time. For example temperature readings, flow of a river, concentration of a chemical process etc. can be recorded as a continuous time series. On the other hand population of a particular city, production of a company, exchange rates between two different currencies may represent discrete time series. Usually in a discrete time series the consecutive observations are recorded at equally spaced time intervals such as hourly, daily, weekly, monthly or yearly time separations. As mentioned in [23], the variable being observed in a discrete time series is assumed to be measured as a continuous variable using the real number scale. Furthermore a continuous time series can be easily transformed to a discrete one by merging data together over a specified time interval.

### 2.2 Components of a Time Series

A time series in general is supposed to be affected by four main components, which can be separated from the observed data. These components are: *Trend, Cyclical, Seasonal* and *Irregular* components. A brief description of these four components is given here.

The general tendency of a time series to increase, decrease or stagnate over a long period of time is termed as Secular Trend or simply Trend. Thus, it can be said that trend is a long term movement in a time series. For example, series relating to population growth, number of houses in a city etc. show upward trend, whereas downward trend can be observed in series relating to mortality rates, epidemics, etc.

Seasonal variations in a time series are fluctuations within a year during the season. The important factors causing seasonal variations are: climate and weather conditions, customs, traditional habits, etc. For example sales of ice-cream increase in summer, sales of woolen cloths increase in winter. Seasonal variation is an important factor for businessmen, shopkeeper and producers for making proper future plans.

The cyclical variation in a time series describes the medium-term changes in the series, caused by circumstances, which repeat in cycles. The duration of a cycle extends over longer period of time, usually two or more years. Most of the economic and financial time series show some kind of cyclical variation. For example a business cycle consists of four phases, viz.

i) Prosperity, ii) Decline, iii) Depression and iv) Recovery.

Schematically a typical business cycle can be shown as below:

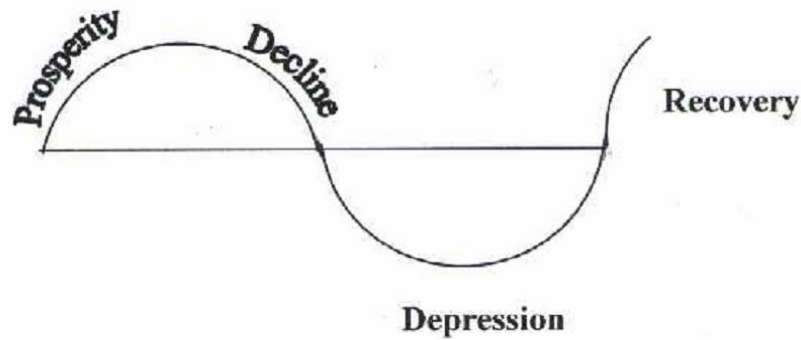

**Fig. 2.1: A four phase business cycle**

Irregular or random variations in a time series are caused by unpredictable influences, which are not regular and also do not repeat in a particular pattern. These variations are caused by incidences such as war, strike, earthquake, flood, revolution, etc. There is no defined statistical technique for measuring random fluctuations in a time series.

Considering the effects of these four components, two different types of models are generally used for a time series viz. Multiplicative and Additive models.

Multiplicative Model: $Y(t) = T(t) \times S(t) \times C(t) \times I(t)$.

Additive Model: $Y(t) = T(t) + S(t) + C(t) + I(t)$.

Here $Y(t)$ is the observation and $T(t), S(t), C(t)$ and $I(t)$ are respectively the trend, seasonal, cyclical and irregular variation at time $t$.

Multiplicative model is based on the assumption that the four components of a time series are not necessarily independent and they can affect one another; whereas in the additive model it is assumed that the four components are independent of each other.

**2.3 Examples of Time Series Data**

Time series observations are frequently encountered in many domains such as business, economics, industry, engineering and science, etc [7, 8, 10]. Depending on the nature of analysis and practical need, there can be various different kinds of time series. To visualize the basic pattern of the data, usually a time series is represented by a graph, where the observations are plotted against corresponding time. Below we show two time series plots:

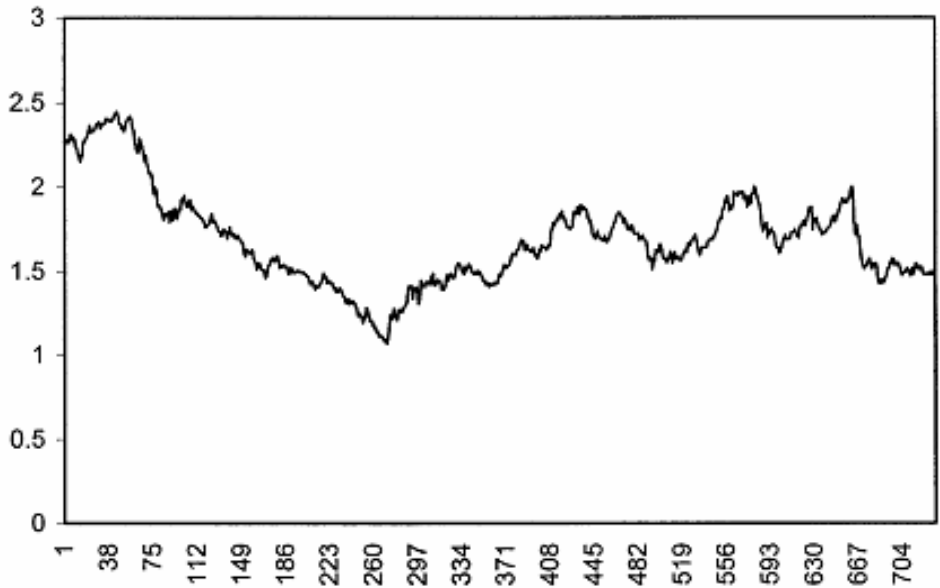

**Fig. 2.2: Weekly BP/USD exchange rate series (1980-1993)**

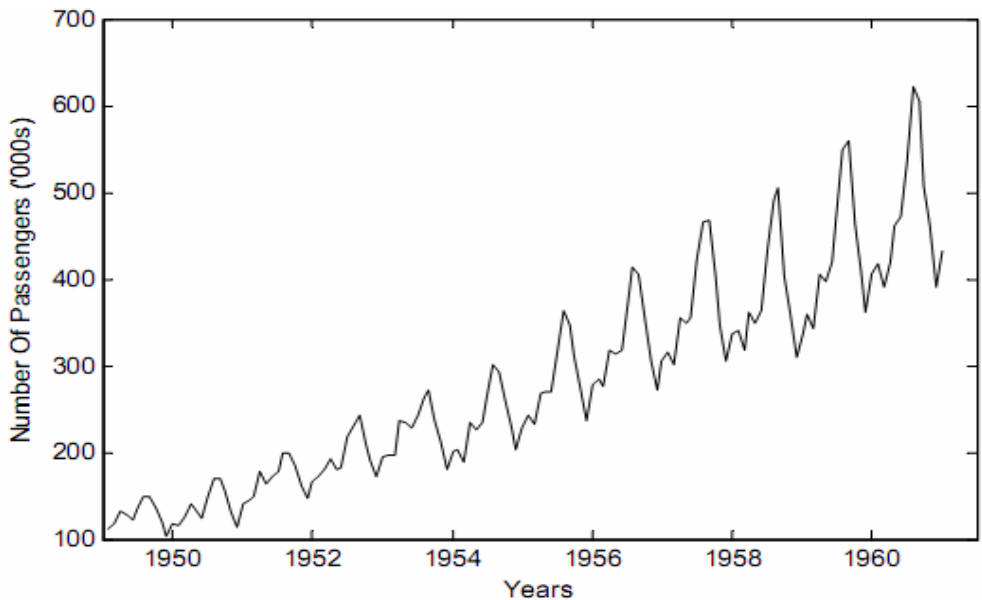

**Fig. 2.3: Monthly international airline passenger series (Jan. 1949-Dec. 1960)**

The first time series is taken from [8] and it represents the weekly exchange rate between British pound and US dollar from 1980 to 1933. The second one is a seasonal time series, considered in [3, 6, 11] and it shows the number of international airline passengers (in thousands) between Jan. 1949 to Dec. 1960 on a monthly basis.

- 14 -

## 2.4 Introduction to Time Series Analysis

In practice a suitable model is fitted to a given time series and the corresponding parameters are estimated using the known data values. The procedure of fitting a time series to a proper model is termed as *Time Series Analysis* [23]. It comprises methods that attempt to understand the nature of the series and is often useful for future forecasting and simulation.

In time series forecasting, past observations are collected and analyzed to develop a suitable mathematical model which captures the underlying data generating process for the series [7, 8]. The future events are then predicted using the model. This approach is particularly useful when there is not much knowledge about the statistical pattern followed by the successive observations or when there is a lack of a satisfactory explanatory model. Time series forecasting has important applications in various fields. Often valuable strategic decisions and precautionary measures are taken based on the forecast results. Thus making a good forecast, i.e. fitting an adequate model to a time series is vary important. Over the past several decades many efforts have been made by researchers for the development and improvement of suitable time series forecasting models.

## 2.5 Time Series and Stochastic Process

A time series is non-deterministic in nature, i.e. we cannot predict with certainty what will occur in future. Generally a time series $\{x(t), t = 0, 1, 2, ...\}$ is assumed to follow certain probability model [21] which describes the joint distribution of the random variable $x_t$. The mathematical expression describing the probability structure of a time series is termed as a stochastic process [23]. Thus the sequence of observations of the series is actually a sample realization of the stochastic process that produced it.

A usual assumption is that the time series variables $x_t$ are independent and identically distributed (i.i.d) following the normal distribution. However as mentioned in [21], an interesting point is that time series are in fact not exactly i.i.d; they follow more or less some regular pattern in long term. For example if the temperature today of a particular city is extremely high, then it can be reasonably presumed that tomorrow's temperature will also likely to be high. This is the reason why time series forecasting using a proper technique, yields result close to the actual value.

## 2.6 Concept of Stationarity

The concept of stationarity of a stochastic process can be visualized as a form of statistical equilibrium [23]. The statistical properties such as mean and variance of a stationary process do not depend upon time. It is a necessary condition for building a time series model that is useful for future forecasting. Further, the mathematical complexity of the fitted model reduces with this assumption. There are two types of stationary processes which are defined below:

A process $\{x(t), t = 0, 1, 2, ...\}$ is *Strongly Stationary* or *Strictly Stationary* if the joint probability distribution function of $\{x_{t-s}, x_{t-s+1}, ..., x_t, ... x_{t+s-1}, x_{t+s}\}$ is independent of $t$ for all $s$. Thus for a strong stationary process the joint distribution of any possible set of random variables from the process is independent of time [21, 23].

However for practical applications, the assumption of strong stationarity is not always needed and so a somewhat weaker form is considered. A stochastic process is said to be *Weakly Stationary* of order $k$ if the statistical moments of the process up to that order depend only on time differences and not upon the time of occurrences of the data being used to estimate the moments [12, 21, 23]. For example a stochastic process $\{x(t), t = 0, 1, 2, ...\}$ is second order stationary [12, 23] if it has time independent mean and variance and the covariance values $Cov(x_t, x_{t-s})$ depend only on $s$.

It is important to note that neither strong nor weak stationarity implies the other. However, a weakly stationary process following normal distribution is also strongly stationary [21]. Some mathematical tests like the one given by Dickey and Fuller [21] are generally used to detect stationarity in a time series data.

As mentioned in [6, 23], the concept of stationarity is a mathematical idea constructed to simplify the theoretical and practical development of stochastic processes. To design a proper model, adequate for future forecasting, the underlying time series is expected to be stationary. Unfortunately it is not always the case. As stated by Hipel and McLeod [23], the greater the time span of historical observations, the greater is the chance that the time series will exhibit non-stationary characteristics. However for relatively short time span, one can reasonably model the series using a stationary stochastic process. Usually time series, showing trend or seasonal patterns are non-stationary in nature. In such cases, differencing and power transformations are often used to remove the trend and to make the series stationary. In the next chapter we shall discuss about the seasonal differencing technique applied to make a seasonal time series stationary.

**2.7 Model Parsimony**

While building a proper time series model we have to consider the principle of parsimony [2, 7, 8, 23]. According to this principle, always the model with smallest possible number of parameters is to be selected so as to provide an adequate representation of the underlying time series data [2]. Out of a number of suitable models, one should consider the simplest one, still maintaining an accurate description of inherent properties of the time series. The idea of model parsimony is similar to the famous *Occam's razor* principle [23]. As discussed by Hipel and McLeod [23], one aspect of this principle is that when face with a number of competing and adequate explanations, pick the most simple one. The Occam's razor provides considerable inherent informations, when applied to logical analysis.

Moreover, the more complicated the model, the more possibilities will arise for departure from the actual model assumptions. With the increase of model parameters, the risk of overfitting also subsequently increases. An over fitted time series model may describe the training data very well, but it may not be suitable for future forecasting. As potential overfitting affects the ability of a model to forecast well, parsimony is often used as a guiding principle to overcome this issue. Thus in summary it can be said that, while making time series forecasts, genuine attention should be given to select the most parsimonious model among all other possibilities.

*Chapter-3*

# Time Series Forecasting Using Stochastic Models

**3.1 Introduction**

In the previous chapter we have discussed about the fundamentals of time series modeling and forecasting. The selection of a proper model is extremely important as it reflects the underlying structure of the series and this fitted model in turn is used for future forecasting. A time series model is said to be linear or non-linear depending on whether the current value of the series is a linear or non-linear function of past observations.

In general models for time series data can have many forms and represent different stochastic processes. There are two widely used linear time series models in literature, viz. *Autoregressive (AR)* [6, 12, 23] and *Moving Average (MA)* [6, 23] models. Combining these two, the *Autoregressive Moving Average (ARMA)* [6, 12, 21, 23] and *Autoregressive Integrated Moving Average (ARIMA)* [6, 21, 23] models have been proposed in literature. The *Autoregressive Fractionally Integrated Moving Average (ARFIMA)* [9, 17] model generalizes ARMA and ARIMA models. For seasonal time series forecasting, a variation of ARIMA, viz. the *Seasonal Autoregressive Integrated Moving Average (SARIMA)* [3, 6, 23] model is used. ARIMA model and its different variations are based on the famous Box-Jenkins principle [6, 8, 12, 23] and so these are also broadly known as the Box-Jenkins models.

Linear models have drawn much attention due to their relative simplicity in understanding and implementation. However many practical time series show non-linear patterns. For example, as mentioned by R. Parrelli in [28], non-linear models are appropriate for predicting volatility changes in economic and financial time series. Considering these facts, various non-linear models have been suggested in literature. Some of them are the famous *Autoregressive Conditional Heteroskedasticity (ARCH)* [9, 28] model and its variations like *Generalized ARCH (GARCH)* [9, 28], *Exponential Generalized ARCH (EGARCH)* [9] etc., the *Threshold Autoregressive (TAR)* [8, 10] model, the *Non-linear Autoregressive (NAR)* [7] model, the *Non-linear Moving Average (NMA)* [28] model, etc.

In this chapter we shall discuss about the important linear and non-linear stochastic time series models with their different properties. This chapter will provide a background for the upcoming chapters, in which we shall study other models used for time series forecasting.

**3.2 The Autoregressive Moving Average (ARMA) Models**

An ARMA($p, q$) model is a combination of AR($p$) and MA($q$) models and is suitable for univariate time series modeling. In an AR($p$) model the future value of a variable is assumed to be a linear combination of $p$ past observations and a random error together with a constant term. Mathematically the AR($p$) model can be expressed as [12, 23]:

$$y_t = c + \sum_{i=1}^{p} \varphi_i y_{t-i} + \varepsilon_t = c + \varphi_1 y_{t-1} + \varphi_2 y_{t-2} + \ldots\ldots\ldots + \varphi_p y_{t-p} + \varepsilon_t \qquad (3.1)$$

Here $y_t$ and $\varepsilon_t$ are respectively the actual value and random error (or random shock) at time period $t$, $\varphi_i$ $(i = 1, 2, \ldots, p)$ are model parameters and $c$ is a constant. The integer constant $p$ is known as the order of the model. Sometimes the constant term is omitted for simplicity. Usually For estimating parameters of an AR process using the given time series, the Yule-Walker equations [23] are used.

Just as an AR($p$) model regress against past values of the series, an MA($q$) model uses past errors as the explanatory variables. The MA($q$) model is given by [12, 21, 23]:

$$y_t = \mu + \sum_{j=1}^{q} \theta_j \varepsilon_{t-j} + \varepsilon_t = \mu + \theta_1 \varepsilon_{t-1} + \theta_2 \varepsilon_{t-2} + \ldots\ldots\ldots + \theta_q \varepsilon_{t-q} + \varepsilon_t \qquad (3.2)$$

Here $\mu$ is the mean of the series, $\theta_j$ $(j = 1, 2, \ldots, q)$ are the model parameters and $q$ is the order of the model. The random shocks are assumed to be a white noise [21, 23] process, i.e. a sequence of independent and identically distributed (i.i.d) random variables with zero mean and a constant variance $\sigma^2$. Generally, the random shocks are assumed to follow the typical normal distribution. Thus conceptually a moving average model is a linear regression of the current observation of the time series against the random shocks of one or more prior observations. Fitting an MA model to a time series is more complicated than fitting an AR model because in the former one the random error terms are not fore-seeable.

Autoregressive (AR) and moving average (MA) models can be effectively combined together to form a general and useful class of time series models, known as the ARMA models. Mathematically an ARMA($p$, $q$) model is represented as [12, 21, 23]:

$$y_t = c + \varepsilon_t + \sum_{i=1}^{p} \varphi_i y_{t-i} + \sum_{j=1}^{q} \theta_j \varepsilon_{t-j} \qquad (3.3)$$

Here the model orders $p, q$ refer to $p$ autoregressive and $q$ moving average terms.

Usually ARMA models are manipulated using the lag operator [21, 23] notation. The lag or backshift operator is defined as $Ly_t = y_{t-1}$. Polynomials of lag operator or lag polynomials are used to represent ARMA models as follows [21]:

AR($p$) model: $\varepsilon_t = \varphi(L) y_t$.

MA($q$) model: $y_t = \theta(L)\varepsilon_t$.

ARMA($p, q$) model: $\varphi(L)y_t = \theta(L)\varepsilon_t$.

Here $\varphi(L) = 1 - \sum_{i=1}^{p}\varphi_i L^i$ and $\theta(L) = 1 + \sum_{j=1}^{q}\theta_j L_j$.

It is shown in [23] that an important property of AR($p$) process is invertibility, i.e. an AR($p$) process can always be written in terms of an MA($\infty$) process. Whereas for an MA($q$) process to be invertible, all the roots of the equation $\theta(L) = 0$ must lie outside the unit circle. This condition is known as the *Invertibility Condition* for an MA process.

### 3.3 Stationarity Analysis

When an AR($p$) process is represented as $\varepsilon_t = \varphi(L)y_t$, then $\varphi(L) = 0$ is known as the characteristic equation for the process. It is proved by Box and Jenkins [6] that a necessary and sufficient condition for the AR($p$) process to be stationary is that all the roots of the characteristic equation must fall outside the unit circle. Hipel and McLeod [23] mentioned another simple algorithm (by Schur and Pagano) for determining stationarity of an AR process. For example as shown in [12] the AR(1) model $y_t = c + \varphi_1 y_{t-1} + \varepsilon_t$ is stationary when $|\varphi_1| < 1$, with a constant mean $\mu = \dfrac{c}{1-\varphi_1}$ and constant variance $\gamma_0 = \dfrac{\sigma^2}{1-\varphi_1^2}$.

An MA($q$) process is always stationary, irrespective of the values the MA parameters [23]. The conditions regarding stationarity and invertibility of AR and MA processes also hold for an ARMA process. An ARMA($p, q$) process is stationary if all the roots of the characteristic equation $\varphi(L) = 0$ lie outside the unit circle. Similarly, if all the roots of the lag equation $\theta(L) = 0$ lie outside the unit circle, then the ARMA($p, q$) process is invertible and can be expressed as a pure AR process.

### 3.4 Autocorrelation and Partial Autocorrelation Functions (ACF and PACF)

To determine a proper model for a given time series data, it is necessary to carry out the ACF and PACF analysis. These statistical measures reflect how the observations in a time series are related to each other. For modeling and forecasting purpose it is often useful to plot the ACF and PACF against consecutive time lags. These plots help in determining the order of AR and MA terms. Below we give their mathematical definitions:

For a time series $\{x(t), t = 0, 1, 2, ...\}$ the *Autocovariance* [21, 23] at lag $k$ is defined as:

$$\gamma_k = Cov(x_t, x_{t+k}) = E[(x_t - \mu)(x_{t+k} - \mu)] \tag{3.4}$$

The *Autocorrelation Coeffient* [21, 23] at lag *k* is defined as:

$$\rho_k = \frac{\gamma_k}{\gamma_0} \qquad (3.5)$$

Here $\mu$ is the mean of the time series, i.e. $\mu = E[x_t]$. The autocovariance at lag zero i.e. $\gamma_0$ is the variance of the time series. From the definition it is clear that the autocorrelation coefficient $\rho_k$ is dimensionless and so is independent of the scale of measurement. Also, clearly $-1 \leq \rho_k \leq 1$. Statisticians Box and Jenkins [6] termed $\gamma_k$ as the theoretical *Autocovariance Function (ACVF)* and $\rho_k$ as the theoretical *Autocorrelation Function (ACF)*.

Another measure, known as the *Partial Autucorrelation Function (PACF)* is used to measure the correlation between an observation *k* period ago and the current observation, after controlling for observations at intermediate lags (i.e. at lags $< k$) [12]. At lag 1, PACF(1) is same as ACF(1). The detailed formulae for calculating PACF are given in [6, 23].

Normally, the stochastic process governing a time series is unknown and so it is not possible to determine the actual or theoretical ACF and PACF values. Rather these values are to be estimated from the training data, i.e. the known time series at hand. The estimated ACF and PACF values from the training data are respectively termed as sample ACF and PACF [6, 23]. As given in [23], the most appropriate sample estimate for the ACVF at lag *k* is

$$c_k = \frac{1}{n}\sum_{t=1}^{n-k}(x_t - \mu)(x_{t+k} - \mu) \qquad (3.6)$$

Then the estimate for the sample ACF at lag *k* is given by

$$r_k = \frac{c_k}{c_0} \qquad (3.7)$$

Here $\{x(t), t = 0,1,2,.......\}$ is the training series of size *n* with mean $\mu$.

As explained by Box and Jenkins [6], the sample ACF plot is useful in determining the type of model to fit to a time series of length *N*. Since ACF is symmetrical about lag zero, it is only required to plot the sample ACF for positive lags, from lag one onwards to a maximum lag of about *N/4*. The sample PACF plot helps in identifying the maximum order of an AR process. The methods for calculating ACF and PACF for ARMA models are described in [23]. We shall demonstrate the use of these plots for our practical datasets in Chapter 7.

**3.5 Autoregressive Integrated Moving Average (ARIMA) Models**

The ARMA models, described above can only be used for stationary time series data. However in practice many time series such as those related to socio-economic [23] and business show non-stationary behavior. Time series, which contain trend and seasonal patterns, are also non-stationary in nature [3, 11]. Thus from application view point ARMA models are inadequate to properly describe non-stationary time series, which are frequently encountered in practice. For this reason the ARIMA model [6, 23, 27] is proposed, which is a generalization of an ARMA model to include the case of non-stationarity as well.

In ARIMA models a non-stationary time series is made stationary by applying finite differencing of the data points. The mathematical formulation of the ARIMA($p,d,q$) model using lag polynomials is given below [23, 27]:

$$\varphi(L)(1-L)^d y_t = \theta(L)\varepsilon_t, \text{ i.e.}$$
$$\left(1 - \sum_{i=1}^{p} \varphi_i L^i\right)(1-L)^d y_t = \left(1 + \sum_{j=1}^{q} \theta_j L^j\right)\varepsilon_t \quad (3.8)$$

- Here, $p$, $d$ and $q$ are integers greater than or equal to zero and refer to the order of the autoregressive, integrated, and moving average parts of the model respectively.
- The integer $d$ controls the level of differencing. Generally $d=1$ is enough in most cases. When $d=0$, then it reduces to an ARMA($p,q$) model.
- An ARIMA($p,0,0$) is nothing but the AR($p$) model and ARIMA($0,0,q$) is the MA($q$) model.
- ARIMA(0,1,0), i.e. $y_t = y_{t-1} + \varepsilon_t$ is a special one and known as the *Random Walk* model [8, 12, 21]. It is widely used for non-stationary data, like economic and stock price series.

A useful generalization of ARIMA models is the Autoregressive Fractionally Integrated Moving Average (ARFIMA) model, which allows non-integer values of the differencing parameter $d$. ARFIMA has useful application in modeling time series with long memory [17]. In this model the expansion of the term $(1-L)^d$ is to be done by using the general binomial theorem. Various contributions have been made by researchers towards the estimation of the general ARFIMA parameters.

**3.6 Seasonal Autoregressive Integrated Moving Average (SARIMA) Models**

The ARIMA model (3.8) is for non-seasonal non-stationary data. Box and Jenkins [6] have generalized this model to deal with seasonality. Their proposed model is known as the Seasonal ARIMA (SARIMA) model. In this model seasonal differencing of appropriate order is used to remove non-stationarity from the series. A first order seasonal difference is the difference between an observation and the corresponding observation from the previous year and is calculated as $z_t = y_t - y_{t-s}$. For monthly time series $s = 12$ and for quarterly time series $s = 4$. This model is generally termed as the SARIMA$(p,d,q) \times (P,D,Q)^s$ model.

The mathematical formulation of a SARIMA$(p,d,q)\times(P,D,Q)^s$ model in terms of lag polynomials is given below [13]:

$$\Phi_P(L^s)\varphi_p(L)(1-L)^d(1-L^s)^D y_t = \Theta_Q(L^s)\theta_q(L)\varepsilon_t,$$
$$\text{i.e. } \Phi_P(L^s)\varphi_p(L)z_t = \Theta_Q(L^s)\theta_q(L)\varepsilon_t.$$
(3.9)

Here $z_t$ is the seasonally differenced series.

### 3.7 Some Nonlinear Time Series Models

So far we have discussed about linear time series models. As mentioned earlier, nonlinear models should also be considered for better time series analysis and forecasting. Campbell, Lo and McKinley (1997) made important contributions towards this direction. According to them almost all non-linear time series can be divided into two branches: one includes models non-linear in mean and other includes models non-linear in variance (heteroskedastic). As an illustrative example, here we present two nonlinear time series models from [28]:

- *Nonlinear Moving Average (NMA) Model*: $y_t = \varepsilon_t + \alpha\varepsilon_{t-1}^2$. This model is non-linear in mean but not in variance.

- *Eagle's (1982) ARCH Model*: $y_t = \varepsilon_t + \alpha\sqrt{\varepsilon_t^2}$. This model is heteroskedastic, i.e. non-linear in variance, but linear in mean. This model has several other variations, like GARCH, EGARCH etc.

### 3.8 Box-Jenkins Methodology

After describing various time series models, the next issue to our concern is how to select an appropriate model that can produce accurate forecast based on a description of historical pattern in the data and how to determine the optimal model orders. Statisticians George Box and Gwilym Jenkins [6] developed a practical approach to build ARIMA model, which best fit to a given time series and also satisfy the parsimony principle. Their concept has fundamental importance on the area of time series analysis and forecasting [8, 27].

The Box-Jenkins methodology does not assume any particular pattern in the historical data of the series to be forecasted. Rather, it uses a three step iterative approach of *model identification, parameter estimation* and *diagnostic checking* to determine the best parsimonious model from a general class of ARIMA models [6, 8, 12, 27]. This three-step process is repeated several times until a satisfactory model is finally selected. Then this model can be used for forecasting future values of the time series.

The Box-Jenkins forecast method is schematically shown in Fig. 3.1:

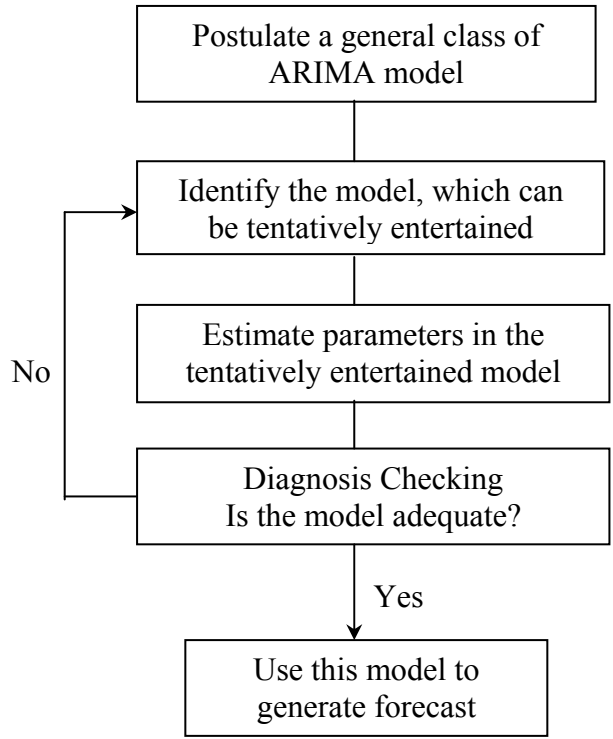

**Fig. 3.1: The Box-Jenkins methodology for optimal model selection**

A crucial step in an appropriate model selection is the determination of optimal model parameters. One criterion is that the sample ACF and PACF, calculated from the training data should match with the corresponding theoretical or actual values [11, 13, 23]. Other widely used measures for model identification are *Akaike Information Criterion (AIC)* [11, 13] and *Bayesian Information Criterion (BIC)* [11, 13] which are defined below [11]:

$$AIC(p) = n\ln(\hat{\sigma}_e^2/n) + 2p$$
$$BIC(p) = n\ln(\hat{\sigma}_e^2/n) + p + p\ln(n)$$

Here *n* is the number of effective observations, used to fit the model, *p* is the number of parameters in the model and $\hat{\sigma}_e^2$ is the sum of sample squared residuals. The optimal model order is chosen by the number of model parameters, which minimizes either AIC or BIC. Other similar criteria have also been proposed in literature for optimal model identification.

*Chapter-4*

# Time Series Forecasting Using Artificial Neural Networks

**4.1 Artificial Neural Networks (ANNs)**

In the previous Chapter we have discussed the important stochastic methods for time series modeling and forecasting. Artificial neural networks (ANNs) approach has been suggested as an alternative technique to time series forecasting and it gained immense popularity in last few years. The basic objective of ANNs was to construct a model for mimicking the intelligence of human brain into machine [13, 20]. Similar to the work of a human brain, ANNs try to recognize regularities and patterns in the input data, learn from experience and then provide generalized results based on their known previous knowledge. Although the development of ANNs was mainly biologically motivated, but afterwards they have been applied in many different areas, especially for forecasting and classification purposes [13, 20]. Below we shall mention the salient features of ANNs, which make them quite favorite for time series analysis and forecasting.

First, ANNs are data-driven and self-adaptive in nature [5, 20]. There is no need to specify a particular model form or to make any *a priori* assumption about the statistical distribution of the data; the desired model is adaptively formed based on the features presented from the data. This approach is quite useful for many practical situations, where no theoretical guidance is available for an appropriate data generation process.

Second, ANNs are inherently non-linear, which makes them more practical and accurate in modeling complex data patterns, as opposed to various traditional linear approaches, such as ARIMA methods [5, 8, 20]. There are many instances, which suggest that ANNs made quite better analysis and forecasting than various linear models.

Finally, as suggested by Hornik and Stinchcombe [22], ANNs are universal functional approximators. They have shown that a network can approximate any continuous function to any desired accuracy [5, 22]. ANNs use parallel processing of the information from the data to approximate a large class of functions with a high degree of accuracy. Further, they can deal with situation, where the input data are erroneous, incomplete or fuzzy [20].

**4.2 The ANN Architecture**

The most widely used ANNs in forecasting problems are multi-layer perceptrons (MLPs), which use a single hidden layer feed forward network (FNN) [5,8]. The model is characterized by a network of three layers, viz. input, hidden and output layer, connected by acyclic links. There may be more than one hidden layer. The nodes in various layers are also known as processing elements. The three-layer feed forward architecture of ANN models can be diagrammatically depicted as below:

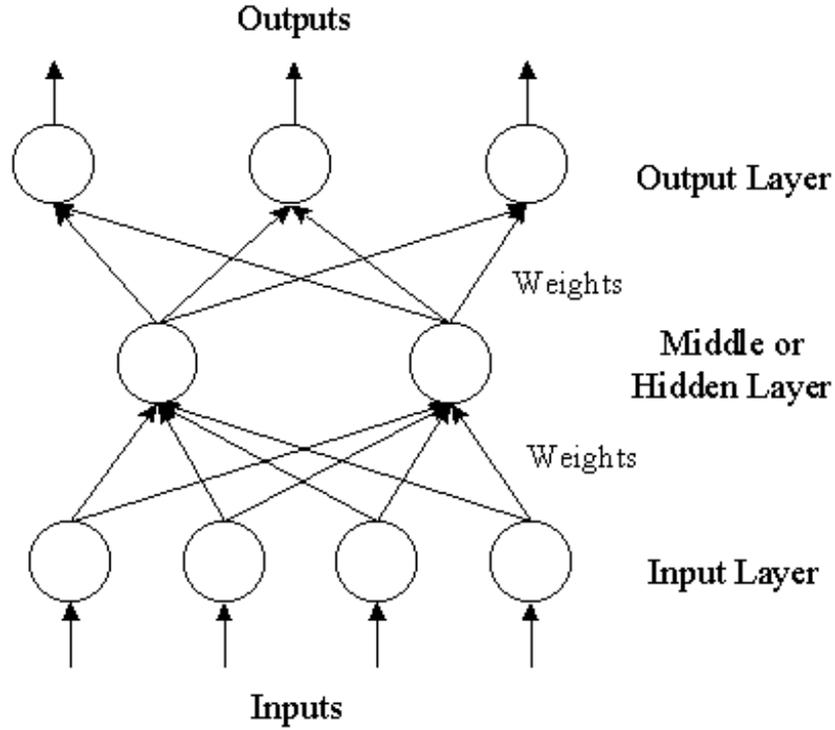

**Fig. 4.1: The three-layer feed forward ANN architecture**

The output of the model is computed using the following mathematical expression [7]:

$$y_t = \alpha_0 + \sum_{j=1}^{q} \alpha_j g\left(\beta_{0j} + \sum_{i=1}^{p} \beta_{ij} y_{t-i}\right) + \varepsilon_t, \forall t \tag{4.1}$$

Here $y_{t-i}(i=1,2,...,p)$ are the $p$ inputs and $y_t$ is the output. The integers $p$, $q$ are the number of input and hidden nodes respectively. $\alpha_j(j=0,1,2,...,q)$ and $\beta_{ij}(i=0,1,2,...,p; j=0,1,2,...,q)$ are the connection weights and $\varepsilon_t$ is the random shock; $\alpha_0$ and $\beta_{0j}$ are the bias terms. Usually, the logistic sigmoid function $g(x) = \dfrac{1}{1+e^{-x}}$ is applied as the nonlinear activation function. Other activation functions, such as linear, hyperbolic tangent, Gaussian, etc. can also be used [20].

The feed forward ANN model (4.1) in fact performs a non-linear functional mapping from the past observations of the time series to the future value, i.e. $y_t = f(y_{t-1}, y_{t-2},.....y_{t-p}, \mathbf{w}) + \varepsilon_t$, where $\mathbf{w}$ is a vector of all parameters and $f$ is a function determined by the network structure and connection weights [5, 8].

To estimate the connection weights, non-linear least square procedures are used, which are based on the minimization of the error function [13]:

$$F(\Psi) = \sum_t e_t^2 = \sum_t (y_t - \hat{y}_t)^2 \tag{4.2}$$

Here $\Psi$ is the space of all connection weights.

The optimization techniques used for minimizing the error function (4.2) are referred as *Learning Rules*. The best-known learning rule in literature is the *Backpropagation* or *Generalized Delta Rule* [13, 20].

### 4.3 Time Lagged Neural Networks (TLNN)

In the FNN formulation, described above, the input nodes are the successive observations of the time series, i.e. the target $x_t$ is a function of the values $x_{t-i}, (i = 1, 2, ..., p)$ where $p$ is the number of input nodes. Another variation of FNN, viz. the TLNN architecture [11, 13] is also widely used. In TLNN, the input nodes are the time series values at some particular lags. For example, a typical TLNN for a time series, with seasonal period $s = 12$ can contain the input nodes as the lagged values at time $t-1$, $t-2$ and $t-12$. The value at time $t$ is to be forecasted using the values at lags 1, 2 and 12.

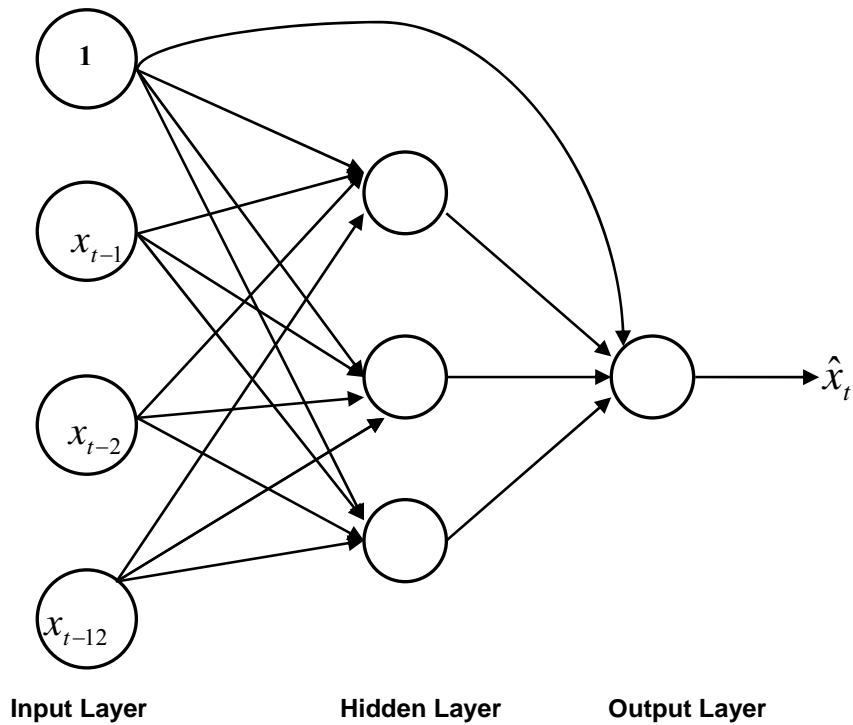

**Fig. 4.2: A typical TLNN architecture for monthly data**

In addition, there is a constant input term, which may be conveniently taken as 1 and this is connected to every neuron in the hidden and output layer. The introduction of this constant input unit avoids the necessity of separately introducing a bias term.

For a TLNN with one hidden level, the general prediction equation for computing a forecast may be written as [11]:

$$\hat{x}_t = \phi_0 + \left\{ w_{c0} + \sum_h w_{h0} \phi_h \left( w_{ch} + \sum_i w_{ih} x_{t-j_i} \right) \right\} \quad (4.3)$$

Here, the selected past observations $x_{t-j_1}, x_{t-j_2}, ..., x_{t-j_k}$ are the input terms, $\{w_{ch}\}$ are the weights for the connections between the constant input and hidden neurons and $w_{c0}$ is the weight of the direct connection between the constant input and the output. Also $\{w_{ih}\}$ and $\{w_{h0}\}$ denote the weights for other connections between the input and hidden neurons and between the hidden and output neurons respectively. $\phi_h$ and $\phi_0$ are the hidden and output layer activation functions respectively.

Faraway and Chatfield [11] used the notation $NN(j_1, j_2, ..., j_k; h)$ to denote the TLNN with inputs at lags $j_1, j_2, ..., j_k$ and $h$ hidden neurons. We shall also adopt this notation in our upcoming experiments. Thus Fig. 4.2 represents an NN (1, 2, 12; 3) model.

**4.4 Seasonal Artificial Neural Networks (SANN)**

The SANN structure is proposed by C. Hamzacebi [3] to improve the forecasting performance of ANNs for seasonal time series data. The proposed SANN model does not require any preprocessing of raw data. Also SANN can learn the seasonal pattern in the series, without removing them, contrary to some other traditional approaches, such as SARIMA, discussed in Chapter 3. The author has empirically verified the good forecasting ability of SANN on four practical time data sets. We have also used this model in our current work on two new seasonal time series and obtained quite satisfactory results. Here we present a brief overview of SANN model as proposed in [3].

In this model, the seasonal parameter $s$ is used to determine the number of input and output neurons. This consideration makes the model surprisingly simple for understanding and implementation. The $i^{th}$ and $(i+1)^{th}$ seasonal period observations are respectively used as the values of input and output neurons in this network structure. Each seasonal period is composed of a number of observations.

Diagrammatically an SANN structure can be shown as [3]:

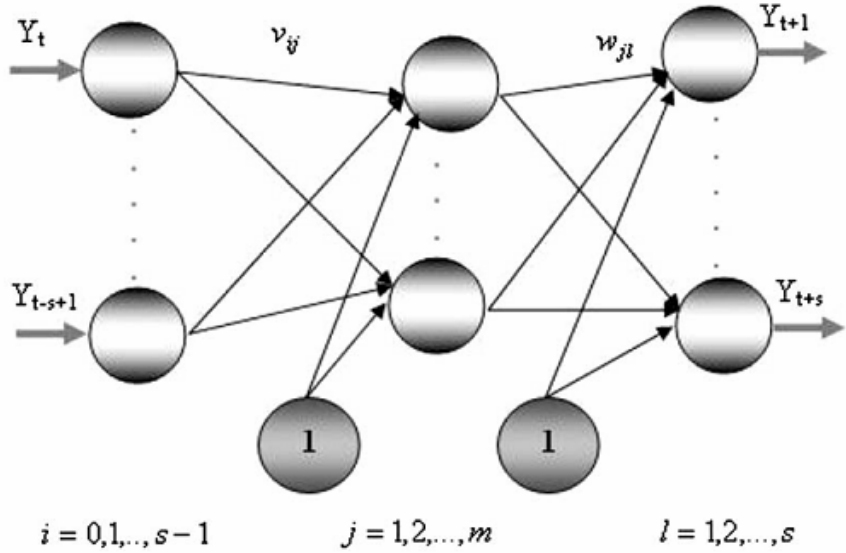

**Fig. 4.3: SANN architecture for seasonal time series**

Mathematical expression for the output of the model is [3]:

$$Y_{t+l} = \alpha_l + \sum_{j=1}^{m} w_{jl} f\left(\theta_j + \sum_{i=0}^{s-1} v_{ij} Y_{t-i}\right) \quad \forall t; l = 1,2,3,...,s. \quad (4.4)$$

Here $Y_{t+l}$ ($l = 1,2,3,...,s$) are the predictions for the future $s$ periods and $Y_{t-i}$ ($i = 0,1,2,...,s-1$) are the observations of the previous $s$ periods; $v_{ij}$ ($i = 0,1,2,...,s-1; j = 1,2,3,...,m$) are weights of connections from input nodes to hidden nodes and $w_{jl}$ ($j = 1,2,3,...,m; l = 1,2,3,...,s$) are weights of connections from hidden nodes to output nodes. Also $\alpha_l$ ($l = 1,2,3,...,s$) and $\theta_j$ ($j = 1,2,3,...,m$) are weights of bias connection and $f$ is the activation function.

Thus while forecasting with SANN, the number of input and output neurons should be taken as 12 for monthly and 4 for quarterly time series. The appropriate number of hidden nodes can be determined by performing suitable experiments on the training data.

### 4.5 Selection of A Proper Network Architecture

So far we have discussed about three important network architectures, viz. the FNN, TLNN and SANN, which are extensively used in forecasting problems. Some other types of neural models are also proposed in literature, such as the *Probabilistic Neural Network (PNN)* [20] for classification problem and *Generalized Regression Neural Network (GRNN)* [20] for regression problem. After specifying a particular network structure, the next most important issue is the determination of the optimal network parameters. The number of network parameters is equal to the total number of connections between the neurons and the bias terms [3, 11].

A desired network model should produce reasonably small error not only on within sample (training) data but also on out of sample (test) data [20]. Due to this reason immense care is required while choosing the number of input and hidden neurons. However, it is a difficult task as there is no theoretical guidance available for the selection of these parameters and often experiments, such as cross-validation are conducted for this purpose [3, 8].

Another major problem is that an inadequate or large number of network parameters may lead to the overtraining of data [2, 11]. Overtraining produces spuriously good within-sample fit, which does not generate better forecasts. To penalize the addition of extra parameters some model comparison criteria, such as AIC and BIC can be used [11, 13]. *Network Pruning* [13] and MacKay's *Bayesian Regularization Algorithm* [11, 20] are also quite popular in this regard.

In summary we can say that NNs are amazingly simple though powerful techniques for time series forecasting. The selection of appropriate network parameters is crucial, while using NN for forecasting purpose. Also a suitable transformation or rescaling of the training data is often necessary to obtain best results.

*Chapter-5*
# Time Series Forecasting Using Support Vector Machines

**5.1 Concept of Support Vector Machines**

Till now, we have studied about various stochastic and neural network methods for time series modeling and forecasting. Despite of their own strengths and weaknesses, these methods are quite successful in forecasting applications. Recently, a new statistical learning theory, viz. the *Support Vector Machine (SVM)* has been receiving increasing attention for classification and forecasting [18, 24, 30, 31]. SVM was developed by Vapnik and his co-workers at the AT & T Bell laboratories in 1995 [24, 29, 33]. Initially SVMs were designed to solve pattern classification problems, such as optimal character recognition, face identification and text classification, etc. But soon they found wide applications in other domains, such as function approximation, regression estimation and time series prediction problems [24, 31, 34].

Vapnik's SVM technique is based on the *Structural Risk Minimization (SRM)* principle [24, 29, 30]. The objective of SVM is to find a decision rule with good generalization ability through selecting some particular subset of training data, called support vectors [29, 31, 33]. In this method, an optimal separating hyperplane is constructed, after nonlinearly mapping the input space into a higher dimensional feature space. Thus, the quality and complexity of SVM solution does not depend directly on the input space [18, 19].

Another important characteristic of SVM is that here the training process is equivalent to solving a linearly constrained quadratic programming problem. So, contrary to other networks' training, the SVM solution is always unique and globally optimal. However a major disadvantage of SVM is that when the training size is large, it requires an enormous amount of computation which increases the time complexity of the solution [24].

Now we are going to present a brief mathematical discussion about SVM concept.

**5.2 Introduction to Statistical Learning Theory**

Vapnik's statistical learning theory is developed in order to derive a learning technique which will provide good generalization. According to Vapnik [33] there are three main problems in machine learning, viz. *Classification*, *Regression* and *Density Estimation*. In all these cases the main goal is to learn a function (or hypothesis) from the training data using a learning machine and then infer general results based on this knowledge.

In case of *supervised learning* the training data is composed of pairs of input and output variables. The input vectors $\mathbf{x} \in X \subseteq \Re^n$ and the output points $y \in D \subseteq \Re$. The two sets $X$ and $D$ are respectively termed as the input space and output space [29, 33]. $D = \{-1, 1\}$ or $\{0, 1\}$ for binary classification problem and $D = \Re$ for regression problem.

In case of *unsupervised learning* the training data is composed of only the input vectors. Here the main goal is to infer the inherent structure of the data through density estimation and clustering technique.

The training data is supposed to be generated from an i.i.d process following an unknown distribution $P(\mathbf{x}, y)$ defined on the set $X \times D$. An input vector is drawn from $X$ with the marginal probability $P(\mathbf{x})$ and the corresponding output point is observed in $D$ with the conditional probability $P(y|\mathbf{x})$.

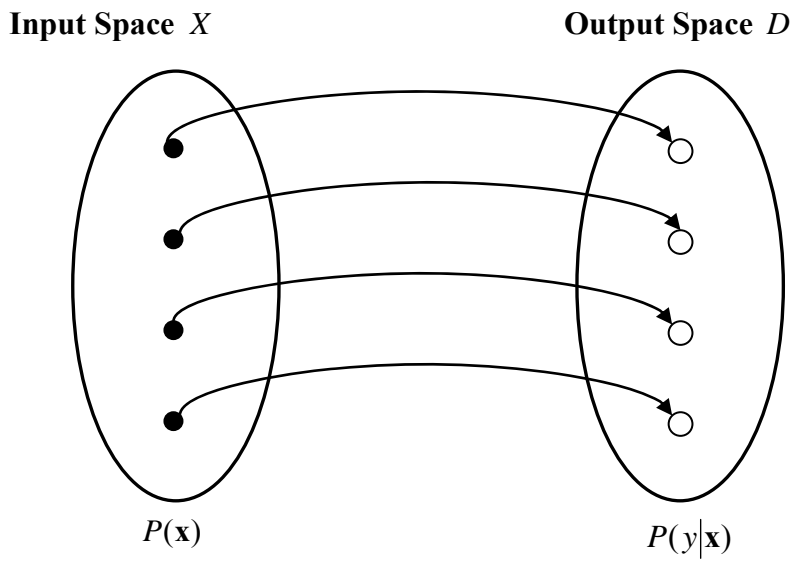

**Fig. 5.1: Probabilistic mapping of input and output points**

After these descriptions, the learning problem can be visualized as searching for the appropriate estimator function $f : X \to D$ which will represent the process of output generations from the input vectors [29, 33]. This function then can be used for generalization, i.e. to produce an output value in response to an unseen input vector.

**5.3 Empirical Risk Minimization (ERM)**

We have seen that the main aim of statistical learning theory is to search for the most appropriate estimator function $f : X \to D$ which maps the points of the input space $X$ to the output space $D$. Following Vapnik and Chervonenkis (1971) first a *Risk Functional* is defined on $X \times D$ to measure the average error occurred among the actual and predicted (or classified) outputs due to using an estimator function $f$. Then the most suitable estimator function is chosen to be that function which minimizes this risk [29, 30, 33].

Let us consider the set of functions $F = \{f(\mathbf{x}, \mathbf{w})\}$ that map the points from the input space $X \subseteq \Re^n$ into the output space $D \subseteq \Re$ where $\mathbf{w}$ denotes the parameters defining $f$. Also suppose that $y$ be the actual output point corresponding to the input vector $\mathbf{x}$. Now if $L(y, f(\mathbf{x}, \mathbf{w}))$ measures the error between the actual value $y$ and the predicted value $f(\mathbf{x}, \mathbf{w})$ for using the prediction function $f$ then the *Expected Risk* is defined as [29, 33]:

$$R(f) = \int L(y, f(\mathbf{x}, \mathbf{w})) dP(\mathbf{x}, y) \qquad (5.1)$$

Here $P(\mathbf{x}, y)$ is the probability distribution followed by the training data. $L(y, f(\mathbf{x}, \mathbf{w}))$ is known as the *Loss Function* and it can be defined in a number of ways [24, 29, 30].

The most suitable prediction function is the one which minimizes the expected risk $R(f)$ and is denoted by $f_0$. This is known as the *Target Function*. The main task of learning problem is now to find out this target function, which is the ideal estimator. Unfortunately this is not possible because the probability distribution $P(\mathbf{x}, y)$ of the given data is unknown and so the expected risk (5.1) cannot be computed. This critical problem motivates Vapnik to suggest the *Empirical Risk Minimization (ERM)* principle [33].

The concept of ERM is to estimate the expected risk $R(f)$ by using the training set. This approximation of $R(f)$ is called the empirical risk. For a given training set $\{\mathbf{x}_i, y_i\}$, where $\mathbf{x}_i \in X \subseteq \Re^n, y_i \in D \subseteq \Re (\forall i = 1,2,3,...,N)$ the empirical risk is defined as [29, 33]:

$$R_{emp}(f) = \frac{1}{N} \sum_{i=1}^{N} L(y_i, f(\mathbf{x}_i, \mathbf{w})) \qquad (5.2)$$

The empirical risk $R_{emp}(f)$ has its own minimizer in $F$, which can be taken as $\hat{f}$. The goal of ERM principle is to approximate the target function $f_0$ by $\hat{f}$. This is possible due to the result that $R(f)$ infact converges to $R_{emp}(f)$ when the training size $N$ is infinitely large [33].

**5.4 Vapnik-Chervonenkis (VC) Dimension**

The VC dimension $h$ of a class of functions $F$ is defined as the maximum number of points that can be exactly classified (i.e. shattered) by $F$ [29, 33]. So mathematically [1, 33]:

$h = \max\{|X|, X \subseteq \Re^n, \text{such that } \forall b \in \{-1,1\}^{|X|}, \exists f \in F \text{ such that } \forall x_i \in X\ (1 \le i \le N), f(x_i) = b_i\}$.

The VC dimension is infact a measure of the intrinsic capacity of a class of functions $F$. It is proved by Burges in 1998 [1] that the VC dimension of the set of oriented hyperplanes in $\Re^n$ is $(n+1)$. Thus three points labeled in eight different ways can always be classified by a linear oriented decision boundary in $\Re^2$ but four points cannot. Thus VC dimension of the set of oriented straight lines in $\Re^2$ is three. For example the XOR problem [29] cannot be realized using a linear decision boundary. However a quadratic decision boundary can correctly classify the points in this problem. This is shown in the figures below:

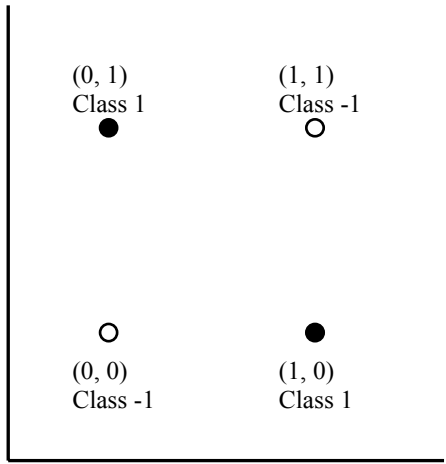 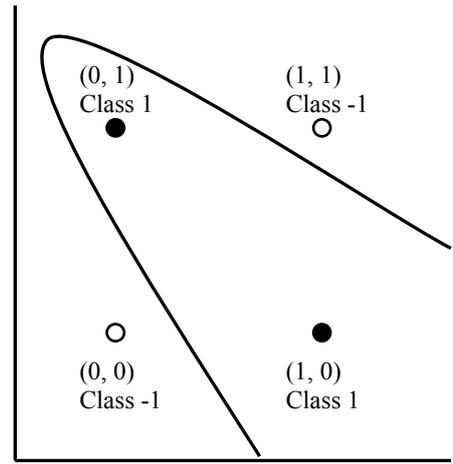

**(a) The linearly non-separable XOR problem**

**(b) A quadratic decision boundary classifying the XOR problem**

**Fig. 5.2: The two-dimensional XOR problem**

**5.5 Structural Risk Minimization (SRM)**

The crucial shortcoming of ERM principle is that in practice we always have a finite set of observations and so it cannot be guaranteed that the estimator function minimizing the empirical risk over $F$ will also minimize the expected risk. To deal with this issue the SRM principle was developed by Vapnik and Chervonenkis in 1982 [33]. The key result motivating this principle is that the difference between the empirical and expected risk can be bounded in terms of the VC dimension of the class $F$ of estimator functions. Below we present the corresponding mathematical theorem for $\{0,1\}$ binary classification problem [29]:

**Theorem:** Let $F$ be a class of estimator functions of VC dimension $h$. Then for any sample $\{\mathbf{x}_i, y_i\}$, where $\mathbf{x}_i \in X \subseteq \Re^n$, $y_i \in D \subseteq \Re$ ($\forall i = 1,2,3,...,N$) drawn from any distribution $P(\mathbf{x}, y)$ the following bound holds true with probability $1 - \eta$ ($0 \leq \eta \leq 1$):

$$R(f) \leq R_{emp}(f) + \sqrt{\frac{h\left(\ln\left(\frac{2N}{h}\right) + 1\right) - \ln\left(\frac{\eta}{4}\right)}{N}} \qquad (5.3)$$

The second term on the right is said to be the *VC Confidence* and $1 - \eta$ is called the *Confidence Level*.

Equation (4.3) is the main inspiration behind the SRM principle. It suggests that to achieve a good generalization one has to minimize the combination of the empirical risk and the complexity of the hypothesis space. In other words one should try to select that hypothesis space which realizes the best trade-off between small empirical error and small model complexity. This concept is similar to the *Bias-Variance Dilemma* of machine learning [29].

## 5.6 Support Vector Machines (SVMs)

The main idea of SVM when applied to binary classification problems is to find a canonical hyperplane which maximally separates the two given classes of training samples [18, 24, 29, 31, 33]. Let us consider two sets of linearly separable training data points in $\Re^n$ which are to be classified into one of the two classes $C_1$ and $C_2$ using linear hyperplanes, (i.e. straight lines). From an infinite number of separating hyperplanes the one with maximum margin is to be selected for best classification and generalization [29, 33]. Below we present a diagrammatic view of this concept:

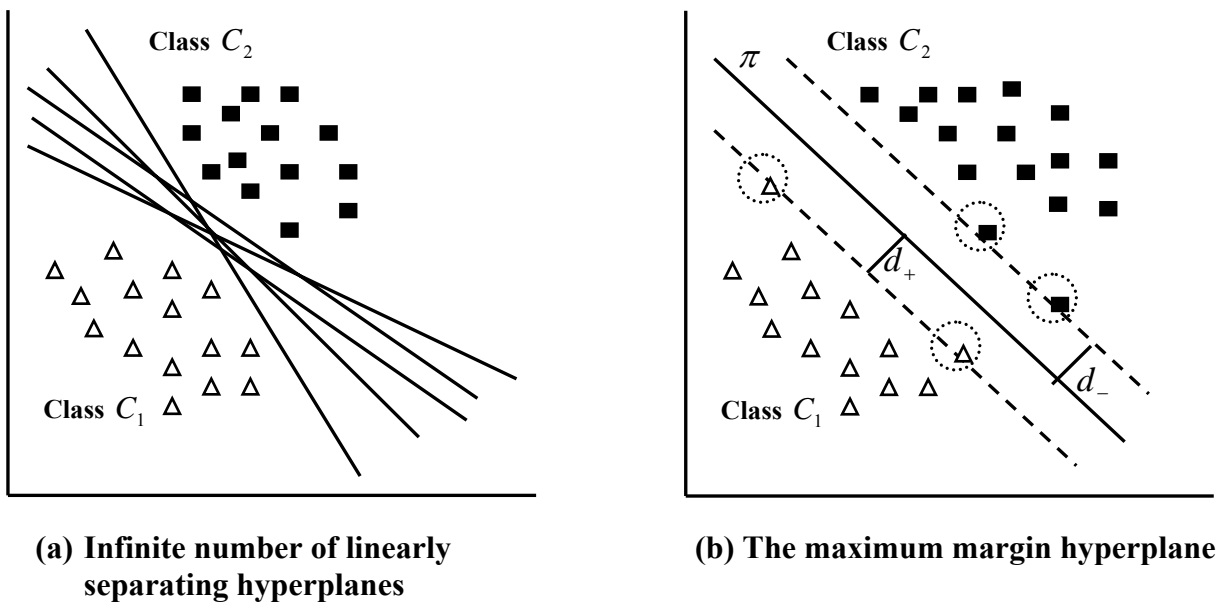

(a) Infinite number of linearly separating hyperplanes

(b) The maximum margin hyperplane

**Fig. 5.3: Support vectors for linearly separable data points**

In Fig.4.3(a) it can be seen that there are an infinite number of hyperplanes, separating the training data points. As shown in Fig.4.3(b), $d_+$ and $d_-$ denote the perpendicular distances from the separating hyperplane to the closest data points of $C_1$ and $C_2$ respectively. Then either of the distances $d_+$ or $d_-$ is termed as the margin and the total margin is $M = d_+ + d_-$. For accurate classification as well as best generalization, the hyperplane which maximizes the total margin is considered as the optimal one and is known as the *Maximum Margin Hyperplane* [29, 33]. Offcourse for this optimal hyperplane $d_+ = d_-$ [29]. The data points from either of the two classes which are closest to the maximum margin hyperplane are known as *Support Vectors* [29, 33]. In Fig.4.3(b) $\pi$ denotes the optimal hyperplane and circulated data points of both the classes represent the support vectors.

Let us consider that the training set is composed of the input-output pairs $\{\mathbf{x}_i, y_i\}_{i=1}^{N}$, where $\mathbf{x}_i \in \Re^n$, $y_i \in \{-1, 1\}$. The goal is to classify the data points into two classes by finding a maximum margin canonical hyperplane. The hypothesis space is the set of functions $f(\mathbf{x}, \mathbf{w}, b) = \text{sgn}(\mathbf{w}^T \mathbf{x} + b)$ where $\mathbf{w}$ is the weight vector, $\mathbf{x} \in \Re^n$ and $b$ is the bias term. The set of separating hyperplanes is given by $\{\mathbf{x} \in \Re^n : \mathbf{w}^T \mathbf{x} + b = 0, \text{where } \mathbf{w} \in \Re^n, b \in \Re\}$. Using SVM the problem of finding the maximum margin hyperplane reduces to solving a non-linear convex quadratic programming problem (QPP) [29, 32, 33].

**SVM for Linearly Separable Data**

For linearly separable data the corresponding quadratic optimization problem is [18, 29, 33]:

$$\left.\begin{array}{l} \text{Minimize} \quad J(\mathbf{w}) = \frac{1}{2}\mathbf{w}^T \mathbf{w} = \frac{1}{2}\|\mathbf{w}\|^2 \\ \text{Subject to} \quad y_i(\mathbf{w}^T \mathbf{x}_i + b) \geq 1; \; \forall i = 1, 2, \ldots, N \end{array}\right\} \quad (5.4)$$

To solve the QPP (5.4) it is conveniently transformed to the dual space. Then Lagrange multipliers and Kühn-Tucker complimentary conditions are used to find the optimal solution. Let us consider that the solution to the QPP yields the optimized Lagrange multipliers $\boldsymbol{\alpha} = (\alpha_1, \alpha_2, \ldots, \alpha_N)^T$, where $\alpha_i \geq 0 \, (\forall i = 1, 2, \ldots, N)$ and the optimal bias $b_{opt}$. The data vectors for which $\alpha_i > 0$ are the support vectors and suppose that there are total $N_s$ support vectors. Then the optimal weight vector can be written as [29, 33]:

$$\mathbf{w}_{opt} = \sum_{i=1}^{N_s} \alpha_i y_i \mathbf{x}_i \quad (5.5)$$

The optimal hyperplane decision function is given by [29, 33]:

$$y(\mathbf{x}) = \text{sgn}\left(\sum_{i=1}^{N_s} y_i \alpha_i (\mathbf{x}^T \mathbf{x}_i) + b_{opt}\right) \quad (5.6)$$

An unknown data is classified in either of the two classes according as the sign of $y(\mathbf{x})$.

**SVM for Non-linearly Separable Data**

In practical applications often the training data points are not linearly separable; as an example we can take the XOR classification problem. In such cases a *Soft Margin Hyperplane* [29] classifier is constructed. The corresponding QPP is given by [18, 29, 33]:

$$\left.\begin{array}{l} \text{Minimize} \quad J(\mathbf{w}, \xi) = \frac{1}{2}\|\mathbf{w}\|^2 + C\left(\sum_{i=1}^{N} \xi_i\right) \\ \text{Subject to} \quad y_i(\mathbf{w}^T \mathbf{x}_i + b) \geq 1 - \xi_i; \; \forall i = 1, 2, \ldots, N \\ \qquad\qquad \xi_i \geq 0 \end{array}\right\} \quad (5.7)$$

Here the slack variables $\xi_i$ are introduced to relax the hard-margin constraints and $C > 0$ is the regularization constant which assigns a penalty to misclassification. Again (4.7) represents a QPP and its solution can be found by applying Lagrange multipliers and Kühn-Tucker conditions. The optimal weight vector and decision function is similar to those in the linearly separable case. The only difference is that in this case the Lagrange multipliers have an upper bound on $C$, i.e. $0 \leq \alpha_i \leq C$ and for support vectors $0 < \alpha_i \leq C$.

**5.7 Support Vector Kernels**

In SVM applications it is convenient to map the points of the input space to a high dimensional *Feature Space* through some non-linear mapping [18, 19, 29] and the optimal separating hyperplane is constructed in this new feature space. This method also resolves the problem where the training points are not separable by a linear decision boundary. Because by using an appropriate transformation the training data points can be made linearly separable in the feature space. A pictorial view of this idea can be obtained from Fig. 4.4:

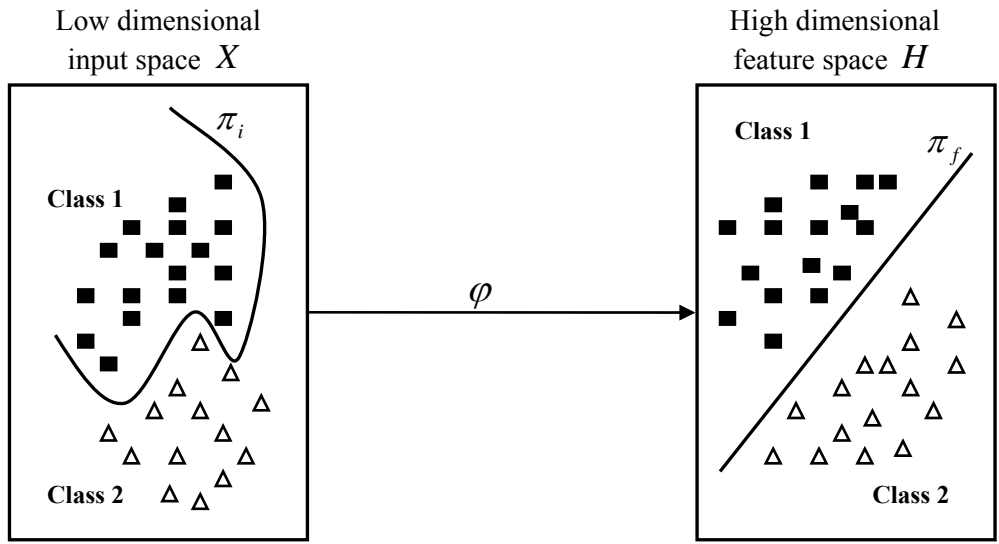

**Fig. 5.4: Non-linear mapping of input space to the feature space**

In Fig. 4.4, the data points of the input space $X \subseteq \Re^n$ are mapped to the feature space $H$ using the nonlinear mapping $\varphi: X \to H$. Due to this transformation, the linearly non-separable input points can now be separated by a linear decision boundary $\pi_f$ in the feature space. The corresponding linear decision function in $H$ can be written as [18, 29, 33]:

$$y(\mathbf{x}) = \operatorname{sgn}\left( \sum_{i=1}^{N_s} y_i \alpha_i \left( \varphi(\mathbf{x})^T \varphi(\mathbf{x}_i) \right) + b_{opt} \right) \tag{5.8}$$

Now if we can find a function $K(\mathbf{x}_i, \mathbf{x}_j) = \varphi(\mathbf{x}_i)^T \varphi(\mathbf{x}_j)$ then it can be directly used in the SVM training equations, without even knowing about the mapping $\varphi(\mathbf{x})$. This function $K$ is known as the *Kernel Function* [18, 29, 30, 34]. To avoid the explicit computation of the non-linear mapping $\varphi(\mathbf{x})$, the associated kernel function must satisfy the *Mercer's Condition* [29, 30, 34]. Below we present some well known kernels used in SVM literature:

- The *Linear Kernel* [34] $K(\mathbf{x}, \mathbf{y}) = \mathbf{x}^T \mathbf{y}$.
- The *Polynomial Kernel* [29, 34] $K(\mathbf{x}, \mathbf{y}) = (1 + \mathbf{x}^T \mathbf{y})^d$.
- The *Radial Basis Function (RBF) Kernel* [29, 34] $K(\mathbf{x}, \mathbf{y}) = \exp(-\|\mathbf{x} - \mathbf{y}\|^2 / 2\sigma^2)$.
- The *Neural Network Kernel* [29] $K(\mathbf{x}, \mathbf{y}) = \tanh(a\mathbf{x}^T \mathbf{y} + b)$ where $a, b$ are constants.

**5.8 SVM for Regression (SVR)**

We have briefly discussed about the SVM techniques used for classification. To use SVM for regression problem Vapnik derived a generalized method [29, 30, 33] which we shall present here in short. Except the outputs $y_i \in \Re$ all other variables will have the same meaning as in case of the classification problems.

In SVR, Vapnik used the $\varepsilon$-*insensitive loss function* defined by [19, 24, 29, 30, 33]:

$$L_\varepsilon(y, f(\mathbf{x}, \mathbf{w})) = \begin{cases} 0 & \text{if } |y - f(\mathbf{x}, \mathbf{w})| \leq \varepsilon \\ |y - f(\mathbf{x}, \mathbf{w})| - \varepsilon & \text{otherwise} \end{cases} \tag{5.9}$$

Then as usual the empirical risk to be minimized is given by:

$$R_{emp}(f) = \frac{1}{N} \sum_{i=1}^{N} L_\varepsilon(y_i, f(\mathbf{x}_i, \mathbf{w})) \tag{5.10}$$

The associated QPP can be written as [18, 23, 28, 29, 32]:

$$\left. \begin{aligned} &\text{Minimize} \quad J(\mathbf{w}, \xi, \xi^*) = \frac{1}{2}\|\mathbf{w}\|^2 + C \sum_{i=1}^{N}(\xi_i + \xi_i^*) \\ &\text{Subject to} \\ &\quad y_i - \mathbf{w}^T \varphi(\mathbf{x}_i) - b \leq \varepsilon + \xi_i; \ \forall i = 1, 2, \ldots, N \\ &\quad \mathbf{w}^T \varphi(\mathbf{x}_i) + b - y_i \leq \varepsilon + \xi_i^* \\ &\quad \xi_i \geq 0 \\ &\quad \xi_i^* \geq 0 \end{aligned} \right\} \tag{5.11}$$

To get the solution of (5.11) two sets of Lagrange multipliers are used which are $\boldsymbol{\alpha} = (\alpha_1, \alpha_2, \ldots, \alpha_N)^T$ and $\boldsymbol{\alpha}^* = (\alpha_1^*, \alpha_2^*, \ldots, \alpha_N^*)^T$ where $0 \leq \alpha_i, \alpha_i^* \leq C$. For support vectors $0 < \alpha_i, \alpha_i^* \leq C$. Finally, the optimal decision hyperplane is obtained as [19, 24, 30, 33]:

$$y(\mathbf{x}) = \sum_{i=1}^{N_s} (\alpha_i - \alpha_i^*) k(\mathbf{x}, \mathbf{x}_i) + b_{opt} \tag{5.12}$$

During the past few years extensive research works have been carried out towards the application of SVM for time series forecasting. As a result many different SVM forecasting algorithms have been derived. Some of them are the *Critical Suppot Vector Machine (CSVM)* [31] algorithm, the *Least Square Support Vector Machine (LS-SVM)* [18] algorithm and its variants, viz. the *Recurrent Least Square Support Vector Machine* [19], the *Dynamic Least Square Support Vector Machine (DLS-SVM)* [34] etc. We are now going to present the celebrated DLS-SVM algorithm developed by Y. Fan et al., 2006 [34] for time series forecasting. Before that we shall give an overview of the LS-SVM technique.

**5.9 The LS-SVM Method**

An LS-SVM formulation employs the equality constraints and a sum-squared error (SSE) cost function, instead of quadratic program in traditional SVM. Consider a training data set of $N$ points $\{\mathbf{x}_i, y_i\}_{i=1}^{N}$ with input data $\mathbf{x}_i \in \Re^n$ and the response $y_i \in \Re$. Then we have the optimization problem [18, 19, 32, 34]:

$$\left.\begin{array}{l} \text{Minimize } \underset{w,b,e}{J(\mathbf{w},e)} = \frac{1}{2}\mathbf{w}^T\mathbf{w} + \frac{1}{2}\gamma\sum_{i=1}^{N} e_i^2 \\ \text{Subject to } y_i = \mathbf{w}^T\varphi(\mathbf{x}_i) + b + e_i; \forall i = 1,2,\ldots,N \end{array}\right\} \quad (5.13)$$

Here $\varphi$ is the non-linear mapping to a higher dimensional space and $\gamma$ is the regularization parameter. The RBF kernel $K(\mathbf{x},\mathbf{y}) = \exp\left(-\|\mathbf{x}-\mathbf{y}\|^2 / 2\sigma^2\right)$ with the tuning parameter $\sigma$ can be employed, which satisfies the Mercer's condition [29, 30, 34]. The primal space model of the optimization problem (5.13) is given by:

$$y = \mathbf{w}^T\varphi(\mathbf{x}) + b \quad (5.14)$$

For computational simplicity and avoiding the case of infinite dimensionality of the weight vector $\mathbf{w}$ the optimization operations are performed in the dual space [18, 19, 34].

The Lagrangian for the problem (5.13) is given by [34]:

$$L(\mathbf{w},b,e;\alpha) = J(w,e) - \sum_{i=1}^{N} \alpha_i \{w^T\varphi(\mathbf{x}_i) + b + e_i - y_i\} \quad (5.15)$$

Here $\boldsymbol{\alpha} = [\alpha_1, \alpha_2, \ldots, \alpha_N]^T$, where $\alpha_i \geq 0 (\forall i = 1,2,\ldots,N)$ are the Lagrange multipliers.

Applying the conditions of optimality, one can compute the partial derivatives of $L$ with respect to $\mathbf{w}, b, e_k, \alpha_k$, equate them to zero and finally eliminating $\mathbf{w}$ and $e_k$ obtain the following linear system of equations [18, 19, 34]:

$$\begin{bmatrix} 0 & \mathbf{1} \\ \mathbf{1}^T & \Omega + \gamma^{-1}\mathbf{I} \end{bmatrix}_{(N+1)\times(N+1)} \begin{bmatrix} b \\ \boldsymbol{\alpha} \end{bmatrix}_{(N+1)\times 1} = \begin{bmatrix} 0 \\ \mathbf{y} \end{bmatrix}_{(N+1)\times 1} \quad (5.16)$$

Here $y = [y_1, y_2, ..., y_N]$, $1 = [1,1,...,1]$ and $\Omega$ with $\Omega(i,j) = k(\mathbf{x}_i, \mathbf{x}_j)(\forall i,j = 1,2,...,N)$ is the kernel matrix. The LS-SVM decision function is thus given by [18, 19, 34]:

$$y(\mathbf{x}) = \sum_{i=1}^{N} \alpha_i k(\mathbf{x}, \mathbf{x}_i) + b \tag{5.17}$$

Here $\boldsymbol{\alpha}, b$ are the solutions of the linear system (5.16).

The main benefit of the LS-SVM technique is that it transforms the traditional QPP to a simultaneous linear system problem, thus ensuring simplicity in computations, fast convergence and high precision [18, 34].

**5.10 The DLS-SVM Method**

DLS-SVM [34] is the modified version of the LS-SVM and is suitable for real time system recognition and time series forecasting. It employs the similar concept but different computation method than the recurrent LS-SVM [19]. The key feature of DLS-SVM is that it can track the dynamics of the nonlinear time-varying systems by deleting one existing data point whenever a new observation is added, thus maintaining a constant window size. Keeping in mind the LS-SVM formulation just discussed, let us consider that [34]:

$$\mathbf{Q}_N = \begin{bmatrix} 0 & \mathbf{1} \\ \mathbf{1}^T & \Omega + \gamma^{-1}\mathbf{I} \end{bmatrix}_{(N+1)\times(N+1)} \tag{5.18}$$

Now for solving (5.16) we need $\mathbf{Q}_N^{-1}$ which is to be computed efficiently. When a new observation is to be added to the existing training set then $\mathbf{Q}_N$ becomes [34]:

$$\mathbf{Q}_{N+1} = \begin{bmatrix} \mathbf{Q}_N & \mathbf{k}_{N+1} \\ \mathbf{k}_{N+1}^T & k_{N+1}^* \end{bmatrix}_{(N+2)\times(N+2)} \tag{5.19}$$

Here $k_{N+1}^* = \gamma^{-1} + k(\mathbf{x}_{N+1}, \mathbf{x}_{N+1}) = \gamma^{-1} + 1$, $\mathbf{k}_{N+1} = [1, k(\mathbf{x}_{N+1}, \mathbf{x}_i)]^T, i = 1,2,...,N$.

To save computation time, matrix inversion lemma is applied to calculate $\mathbf{Q}_{N+1}^{-1}$ as [34]:

$$\mathbf{Q}_{N+1}^{-1} = \begin{bmatrix} \mathbf{Q}_N & \mathbf{k}_{N+1} \\ \mathbf{k}_{N+1}^T & k_{N+1}^* \end{bmatrix} = \begin{bmatrix} \mathbf{Q}_N^{-1} + \mathbf{Q}_N^{-1}\mathbf{k}_{N+1}\mathbf{k}_{N+1}^T \mathbf{Q}_N^{-1}\rho^{-1} & -\mathbf{Q}_N^{-1}\mathbf{k}_{N+1}\rho^{-1} \\ -\mathbf{k}_{N+1}^T \mathbf{Q}_N^{-1}\rho^{-1} & \rho^{-1} \end{bmatrix} \tag{5.20}$$

Here $\rho = k_{N+1}^* - \mathbf{k}_{N+1}^T \mathbf{Q}_N^{-1} \mathbf{k}_{N+1}$. With the above recursion equation, all direct matrix inversions are eliminated. To use the relation (5.20) we only need to know $\mathbf{Q}_N^{-1}$, which is obtained during the solution of (5.16).

When a new data point is added while forecasting then *Pruning* [34] is applied to get rid of the first point and replace it with the new input vector. To remove the first point from the training dataset it is assumed that the new data point has just been added and $\mathbf{Q}_{N+1}^{-1}$ is already known. Now rearrange the training dataset as $[\mathbf{x}_2, \mathbf{x}_3, ..., \mathbf{x}_{N+1}, \mathbf{x}_1]$ and get the matrix:

$$\widehat{\mathbf{Q}}_{N+1} = \begin{bmatrix} \widehat{\mathbf{Q}}_N & \mathbf{k}_1 \\ \mathbf{k}_1^T & k_1^* \end{bmatrix} \tag{5.21}$$

Here $k_1^* = \gamma^{-1} + k(\mathbf{x}_1, \mathbf{x}_1) = \gamma^{-1} + 1$, $\mathbf{k}_1 = [1, k(\mathbf{x}_1, \mathbf{x}_i)]^T$, $i = 1, 2, ..., N+1$.

$\widehat{\mathbf{Q}}_N = \begin{bmatrix} 0 & \mathbf{1}^T \\ \mathbf{1} & \mathbf{\Omega} + \gamma^{-1}\mathbf{I} \end{bmatrix}_{(N+1)\times(N+1)}$ with $\Omega_{ij} = k(\mathbf{x}_i, \mathbf{x}_j), i, j = 2, 3, ..., N+1$.

We notice that the difference between $\mathbf{Q}_{N+1}$ and $\widehat{\mathbf{Q}}_{N+1}$ is only the order of rows and columns and so is $\mathbf{Q}_{N+1}^{-1}$ and $\widehat{\mathbf{Q}}_{N+1}^{-1}$. Thus adjusting the element positions of $\mathbf{Q}_{N+1}^{-1}$ the matrix $\widehat{\mathbf{Q}}_{N+1}^{-1}$ can be obtained and then again by matrix inversion lemma [34]:

$$\widehat{\mathbf{Q}}_{N+1}^{-1} = \begin{bmatrix} \widehat{\mathbf{Q}}_N & \mathbf{k}_1 \\ \mathbf{k}_1^T & k_1^* \end{bmatrix}^{-1} = \begin{bmatrix} \mathbf{Q}^*_{(N+1)\times(N+1)} & \mathbf{P}_{(N+1)\times 1} \\ \mathbf{P}^T & q \end{bmatrix} \tag{5.22}$$

Then using (5.20) and (5.22), one can compute [34]:

$$\widehat{\mathbf{Q}}_N^{-1} = \mathbf{Q}^* - \frac{\mathbf{P}\mathbf{P}^T}{q} \tag{5.23}$$

Now we can again compute $\boldsymbol{\alpha}$ and $b$ from (5.16) with $\widehat{\mathbf{Q}}_N^{-1}$ and repeat the steps of pruning until all the data points are exhausted. In this way the DLS-SVM method can be applied for real time series forecasting with efficiency and reduced computation time. A working algorithm for using DLS-SVM for forecasting is given in [34].

We shall now conclude this chapter after touching an important point. The success of SVM to produce a close forecast depends a lot on the proper selection of the hyper-parameters such as the kernel parameter $\sigma$, the regularization constant $\gamma$, the SVR constant $\varepsilon$, etc. An improper choice of these parameters may result in totally ridiculous forecast. As there is no structured way to choose the best hyper-parameters in advance so in practical applications, techniques like cross-validation [24, 34] or Bayesian inference [34] are used. However it should be noted that the values of the optimal hyper-parameters selected in advance could vary afterwards due to different characteristics of future observations [24].

*Chapter-6*

# Forecast Performance Measures

**6.1 Making Real Time Forecasts: A Few Points**

In the previous three chapters, we have studied various useful and popular techniques for time series forecasting. The next important issue is offcourse implementation, i.e. to apply these methods for generating forecasts. While applying a particular model to some real or simulated time series, first the raw data is divided into two parts, viz. the *Training Set* and *Test Set*. The observations in the training set are used for constructing the desired model. Often a small subpart of the training set is kept for validation purpose and is known as the *Validation Set*. Sometimes a preprocessing is done by normalizing the data or taking logarithmic or other transforms. One such famous technique is the *Box-Cox Transformation* [23]. Once a model is constructed, it is used for generating forecasts. The test set observations are kept for verifying how accurate the fitted model performed in forecasting these values. If necessary, an inverse transformation is applied on the forecasted values to convert them in original scale. In order to judge the forecasting accuracy of a particular model or for evaluating and comparing different models, their relative performance on the test dataset is considered.

Due to the fundamental importance of time series forecasting in many practical situations, proper care should be taken while selecting a particular model. For this reason, various performance measures are proposed in literature [3, 7, 8, 9, 24, 27] to estimate forecast accuracy and to compare different models. These are also known as performance metrics [24]. Each of these measures is a function of the actual and forecasted values of the time series.

In this chapter we shall describe few important performance measures which are frequently used by researchers, with their salient features.

**6.2 Description of Various Forecast Performance Measures**

Now we shall discuss about the commonly used performance measures and their important properties. In each of the forthcoming definitions, $y_t$ is the actual value, $f_t$ is the forecasted value, $e_t = y_t - f_t$ is the forecast error and $n$ is the size of the test set. Also, $\bar{y} = \frac{1}{n}\sum_{t=1}^{n} y_t$ is the test mean and $\sigma^2 = \frac{1}{n-1}\sum_{t=1}^{n}(y_t - \bar{y})^2$ is the test variance.

**6.2.1 The Mean Forecast Error (MFE)**

This measure is defined as [24] MFE = $\frac{1}{n}\sum_{t=1}^{n} e_t$. The properties of MFE are:

- It is a measure of the average deviation of forecasted values from actual ones.
- It shows the direction of error and thus also termed as the *Forecast Bias*.
- In MFE, the effects of positive and negative errors cancel out and there is no way to know their exact amount.
- A zero MFE does not mean that forecasts are perfect, i.e. contain no error; rather it only indicates that forecasts are on proper target.
- MFE does not panelize extreme errors.
- It depends on the scale of measurement and also affected by data transformations.
- For a good forecast, i.e. to have a minimum bias, it is desirable that the MFE is as close to zero as possible.

### 6.2.2 The Mean Absolute Error (MAE)

The mean absolute error is defined as [3, 7, 9, 24] MAE $= \frac{1}{n}\sum_{t=1}^{n}|e_t|$. Its properties are:

- It measures the average absolute deviation of forecasted values from original ones.
- It is also termed as the *Mean Absolute Deviation (MAD)*.
- It shows the magnitude of overall error, occurred due to forecasting.
- In MAE, the effects of positive and negative errors do not cancel out.
- Unlike MFE, MAE does not provide any idea about the direction of errors.
- For a good forecast, the obtained MAE should be as small as possible.
- Like MFE, MAE also depends on the scale of measurement and data transformations.
- Extreme forecast errors are not panelized by MAE.

### 6.2.3 The Mean Absolute Percentage Error (MAPE)

This measure is given by [3, 24] MAPE $= \frac{1}{n}\sum_{t=1}^{n}\left|\frac{e_t}{y_t}\right|\times 100$. Its important features are:

- This measure represents the percentage of average absolute error occurred.
- It is independent of the scale of measurement, but affected by data transformation.
- It does not show the direction of error.
- MAPE does not panelize extreme deviations.
- In this measure, opposite signed errors do not offset each other.

### 6.2.4 The Mean Percentage Error (MPE)

It is defined as [27] MPE $= \frac{1}{n}\sum_{t=1}^{n}\left(\frac{e_t}{y_t}\right)\times 100$. The properties of MPE are:

- MPE represents the percentage of average error occurred, while forecasting.
- It has similar properties as MAPE, except,
- It shows the direction of error occurred.
- Opposite signed errors affect each other and cancel out.
- Thus like MFE, by obtaining a value of MPE close to zero, we cannot conclude that the corresponding model performed very well.
- It is desirable that for a good forecast the obtained MPE should be small.

**6.2.5 The Mean Squared Error (MSE)**

Mathematical definition of this measure is [3, 8] MSE = $\frac{1}{n}\sum_{t=1}^{n} e_t^2$. Its properties are:

- It is a measure of average squared deviation of forecasted values.
- As here the opposite signed errors do not offset one another, MSE gives an overall idea of the error occurred during forecasting.
- It panelizes extreme errors occurred while forecasting.
- MSE emphasizes the fact that the total forecast error is in fact much affected by large individual errors, i.e. large errors are much expensive than small errors.
- MSE does not provide any idea about the direction of overall error.
- MSE is sensitive to the change of scale and data transformations.
- Although MSE is a good measure of overall forecast error, but it is not as intuitive and easily interpretable as the other measures discussed before.

**6.2.6 The Sum of Squared Error (SSE)**

It is mathematically defined as [3] SSE = $\sum_{t=1}^{n} e_t^2$.

- It measures the total squared deviation of forecasted observations, from the actual values.
- The properties of SSE are same as those of MSE.

**6.2.7 The Signed Mean Squared Error (SMSE)**

This measure is defined as [24] SMSE = $\frac{1}{n}\sum_{t=1}^{n}\left(\frac{e_t}{|e_t|}\right)e_t^2$. Its salient features are:

- It is same as MSE, except that here the original sign is kept for each individual squared error.
- SMSE panelizes extreme errors, occurred while forecasting.
- Unlike MSE, SMSE also shows the direction of the overall error.

- In calculation of SMSE, positive and negative errors offset each other.
- Like MSE, SMSE is also sensitive to the change of scale and data transformations.

### 6.2.8 The Root Mean Squared Error (RMSE)

Mathematically, RMSE [7, 9] $= \sqrt{\text{MSE}} = \sqrt{\dfrac{1}{n}\sum_{t=1}^{n} e_t^2}$ .

- RMSE is nothing but the square root of calculated MSE.
- All the properties of MSE hold for RMSE as well.

### 6.2.9 The Normalized Mean Squared Error (NMSE)

This measure is defined as [24] $\text{NMSE} = \dfrac{\text{MSE}}{\sigma^2} = \dfrac{1}{\sigma^2 n}\sum_{t=1}^{n} e_t^2$ . Its features are:

- NMSE normalizes the obtained MSE after dividing it by the test variance.
- It is a balanced error measure and is very effective in judging forecast accuracy of a model.
- The smaller the NMSE value, the better forecast.
- Other properties of NMSE are same as those of MSE.

### 6.2.10 The Theil's U-statistics

This important measure is defined as [9] $U = \dfrac{\sqrt{\dfrac{1}{n}\sum_{t=1}^{n} e_t^2}}{\sqrt{\dfrac{1}{n}\sum_{t=1}^{n} f_t^2}\sqrt{\dfrac{1}{n}\sum_{t=1}^{n} y_t^2}}$. Its properties are:

- It is a normalized measure of total forecast error.
- $0 \leq U \leq 1$; $U = 0$ means a perfect fit.
- This measure is affected by change of scale and data transformations.
- For assessing good forecast accuracy, it is desirable that the U-statistic is close to zero.

We have discussed ten important measures for judging forecast accuracy of a fitted model. Each of these measures has some unique properties, different from others. In experiments, it is better to consider more than one performance criteria. This will help to obtain a reasonable knowledge about the amount, magnitude and direction of overall forecast error. For this reason, time series analysts usually use more than one measure for judgment. The next chapter will demonstrate the application of some of these performance measures, in our experimental results.

*Chapter-7*

# Experimental Results

### 7.1 A Brief Overview

Perhaps this is the chapter for which the reader is most eagerly waiting. After gaining a reasonable knowledge about time series modeling and forecasting from the previous chapters, we are now going to implement them on practical datasets.

In this current book, till now we have considered six time series, taken from various sources and research works. All the associated programs are written in MATLAB. To judge forecast performances of different methods, the measures MAD, MSE, RMSE, MAPE and Theil's U-statistics are considered. For each dataset, we have presented our obtained results in tabular form. Also in this chapter we have used the term *Forecast Diagram* to mean the graph showing the test (actual) and forecasted data points. In each forecast diagram, the solid and dotted line respectively represents the actual and forecasted observations.

In all the experiments we have used the RBF kernel for SVM training. Crossvalidation is applied to determine the optimum values of $\sigma$, $C$, $n$ and $N$, where $n$ and $N$ are respectively the dimension and number of input vectors with $(n+N)$ being the size of the training set. The experiments involving all the techniques except SVM are iterated a large number of times. Also rescaling and data transformations are used to some of the datasets.

### 7.2 The Canadian Lynx Dataset

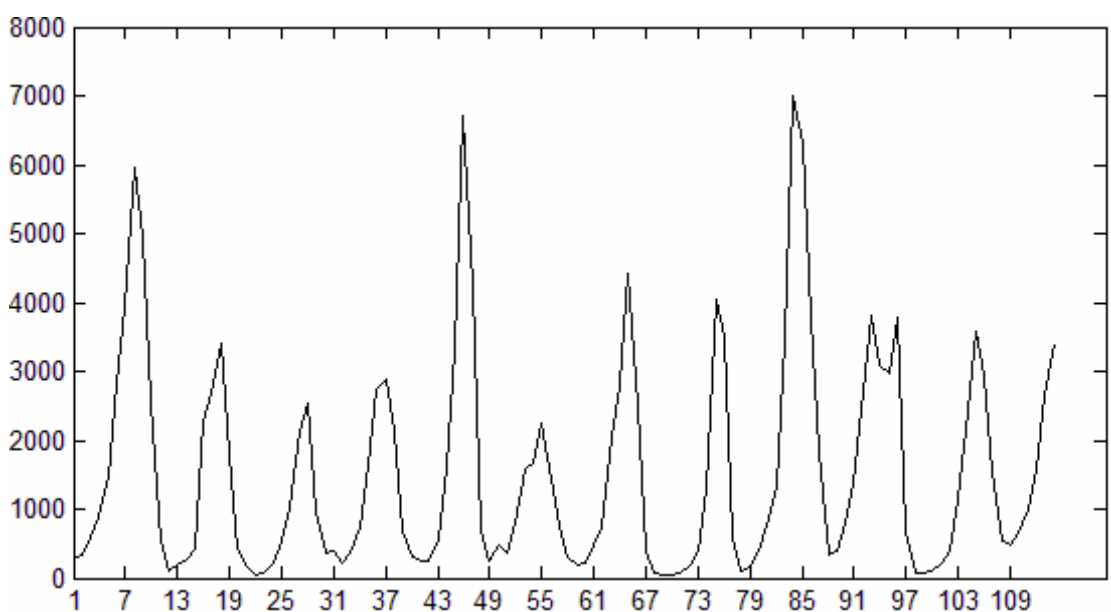

**Fig. 7.2.1: Canadian lynx data series (1821-1934)**

The lynx series shown in Fig. 7.2.1 contains the number of lynx trapped per year in the MacKenzie River district of Northern Canada from 1821 to 1934. It has been extensively studied by time series analysts like G. P. Zhang [8] and M. J. Campbell et al. [26].

The lynx dataset has total 114 observations. Out of these the first 100 are considered for training and the remaining 14 for testing. An AR model of order 12 has been found to be the most parsimonious ARIMA model for this series [8]. Below we have shown the sample ACF and PACF plots for the lynx data series.

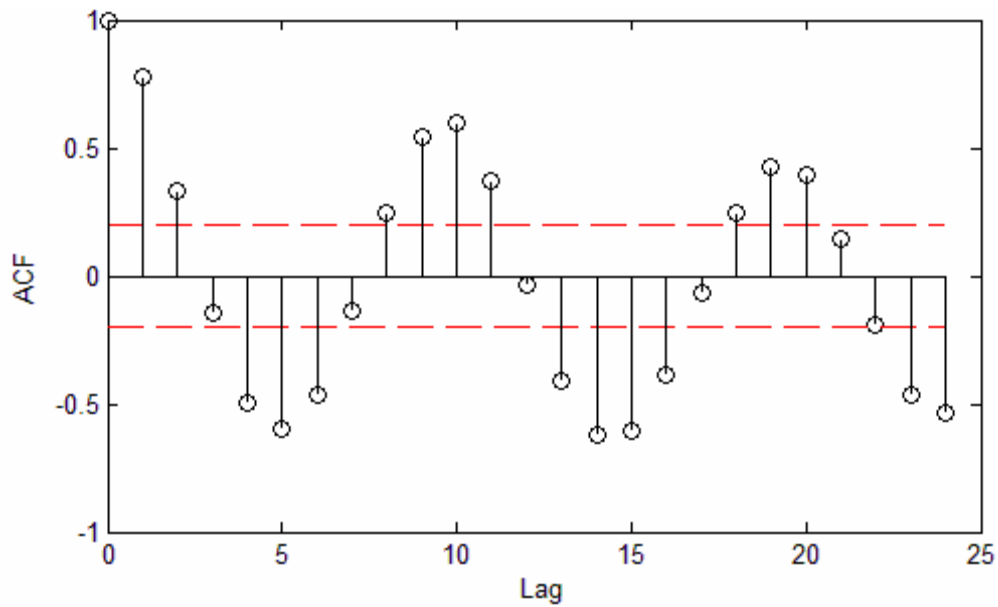

**Fig. 7.2.2: Sample ACF plot for lynx series**

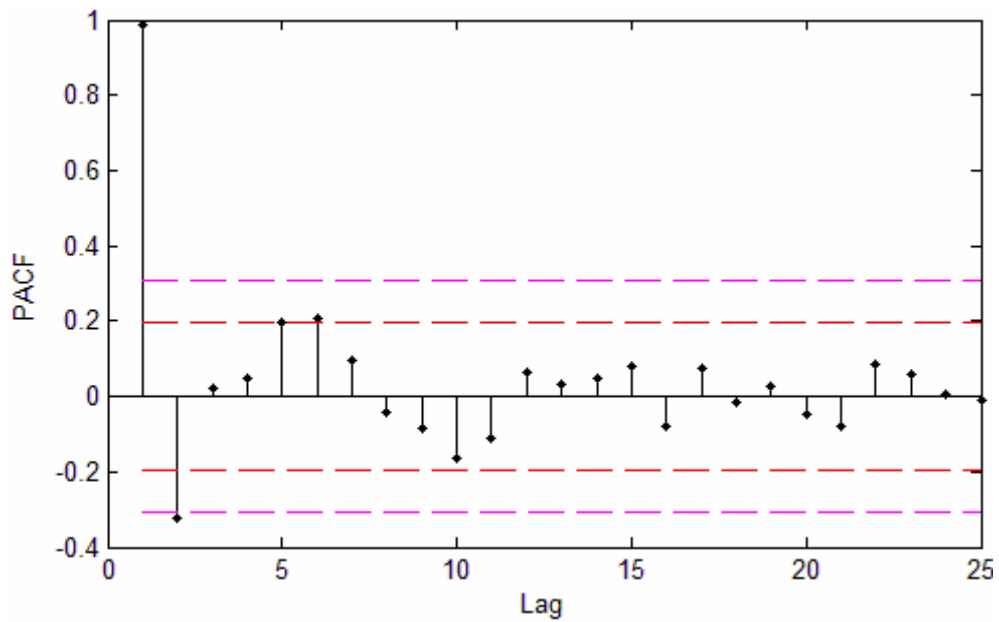

**Fig. 7.2.3: Sample PACF plot for lynx Series**

Following Hipel and McLeod [23], we have considered the ACF and PACF values up to lag 25, i.e. one-fourth of the training size and 95% confidence level. From these plots we can determine the type and order of the adequate model required to fit the series. Moreover as the ACF values attenuate rapidly for increasing lags, we can assume that the lynx series is stationary. Later this fact is also justified by the unit root test.

Similar to [8, 26] we have transformed the lynx series using logarithms to the base 10. Then we have employed AR(12), ARMA(12, 9), ANN and SVM to the series. Following G. P. Zhang [8], a $7 \times 5 \times 1$ ANN structure is used. Though the fitted ARMA(12, 9) model is not parsimonious, but still it generates good forecasts, as can be seen from Table 7.2. Also in this table, for SVM we have given in bracket the optimal hyper-parameter values found by crossvalidation.

The performance measures obtained for the lynx series are shown in Table 7.2:

**Table 7.2: Forecast results for Canadian lynx time series**

| Method | MSE | MAD | RMSE | MAPE | Theil's U Statistics |
|---|---|---|---|---|---|
| AR(12) | 0.005123 | 0.058614 | 0.071577 | 1.950160% | 0.007479 |
| ARMA(12,9) | 0.016533 | 0.096895 | 0.128581 | 3.409039% | 0.013402 |
| ANN | 0.012659 | 0.066743 | 0.112512 | 2.392407% | 0.017836 |
| SVM ($\sigma = 0.8493, C = 1.4126, n = 3, N = 97$) | 0.052676 | 0.173318 | 0.229513 | 5.811812% | 0.023986 |

It can be seen from the above table that the best performance is obtained by using AR(12) model. The four forecast diagrams for lynx series are shown in Fig. 7.2.4.

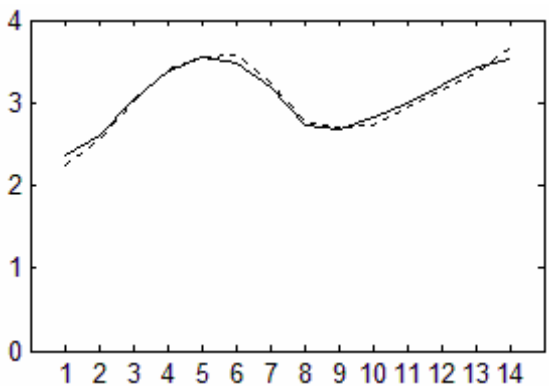
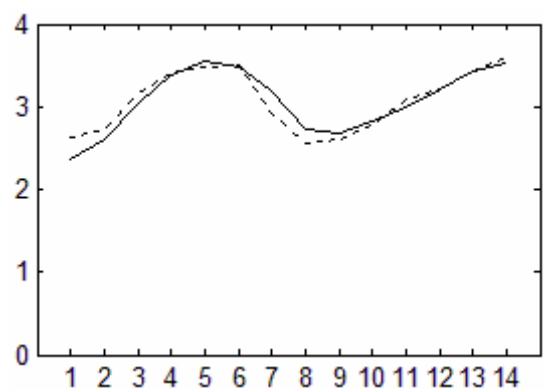

(a) AR(12) forecast diagram           (b) ARMA(12, 9) forecast diagram

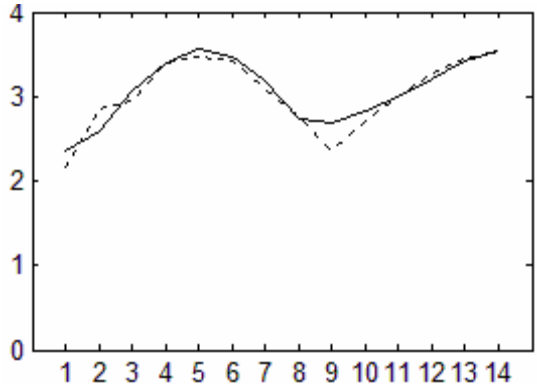 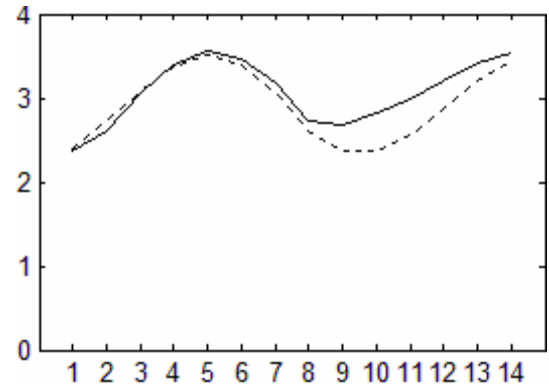

(c) ANN forecast diagram          (d) SVM forecast diagram

Fig. 7.2.4: Forecast diagrams for lynx series

The four forecast diagrams presented in Fig. 7.2.4 are corresponding to the transformed lynx series. The good forecasting performance of the fitted AR(12) model can be visualized from Fig. 7.2.4 (a) as the two graphs are very close to each other.

**7.3 The Wolf's Sunspot Dataset**

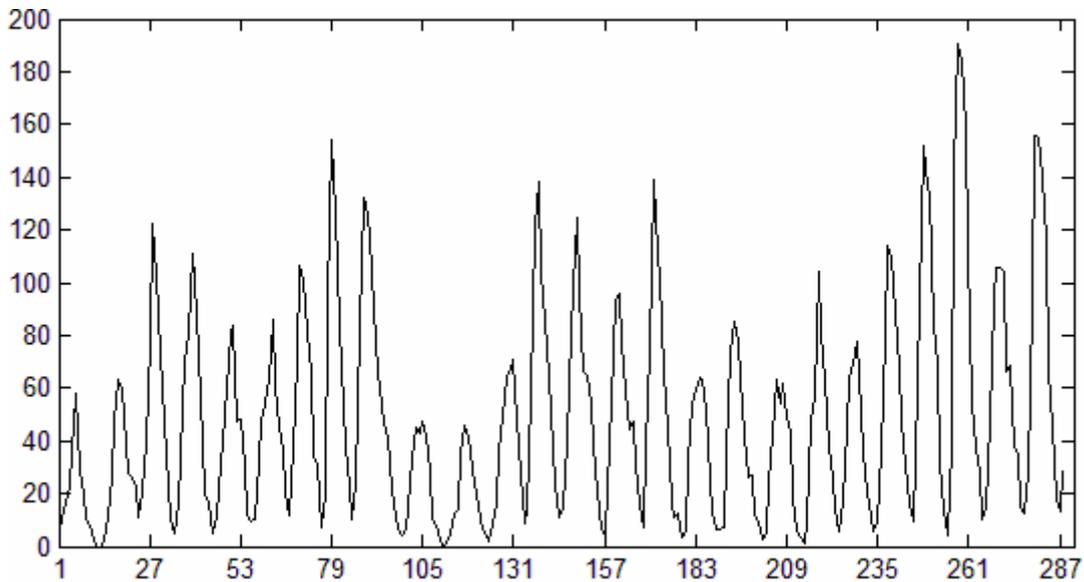

Fig. 7.3.1: Wolf's sunspot data series (1700-1987)

The Wolf's sunspot series shown in Fig. 7.3.1 represents the annual number of sunspots from 1700 to 1987 [8]. This data series is considered as non-linear and it has important practical applications in many domains, such as geophysics, environmental science and climatology [8]. From Fig. 7.3.1 it can be seen that the sunspot series has a mean cycle of about 11 years.

The sunspot dataset has a total of 288 observations, from which the first 221 (i.e. 1700 to 1920) are used for training and the remaining 67 (i.e. 1921 to 1987) for testing. An AR(9) model has been found to be the most parsimonious ARIMA model for this series [8]. The sample ACF and PACF plots with 95% confidence level for the sunspot series are shown in the two figures below:

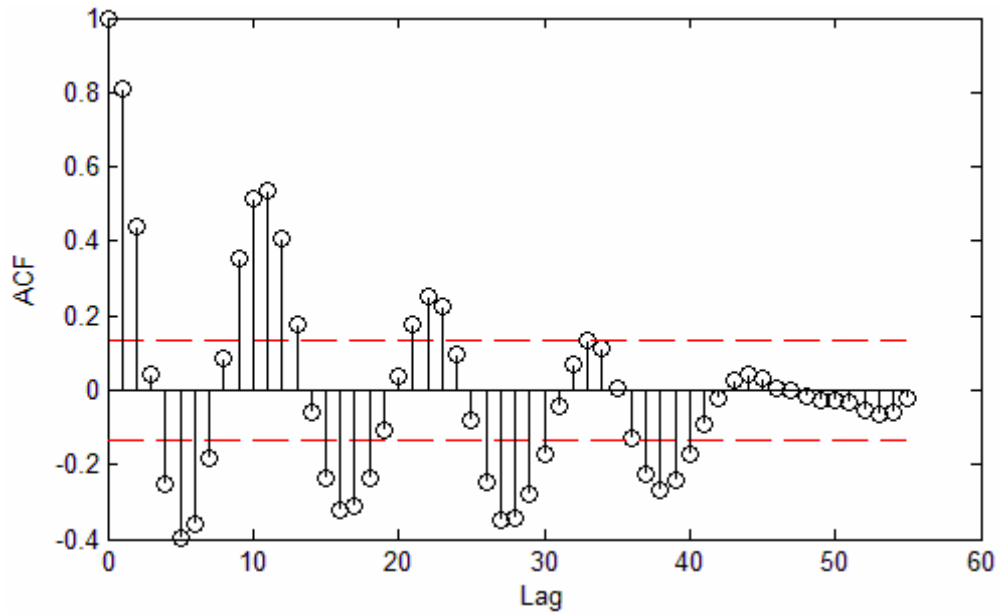

**Fig. 7.3.2: Sample ACF plot for sunspot series**

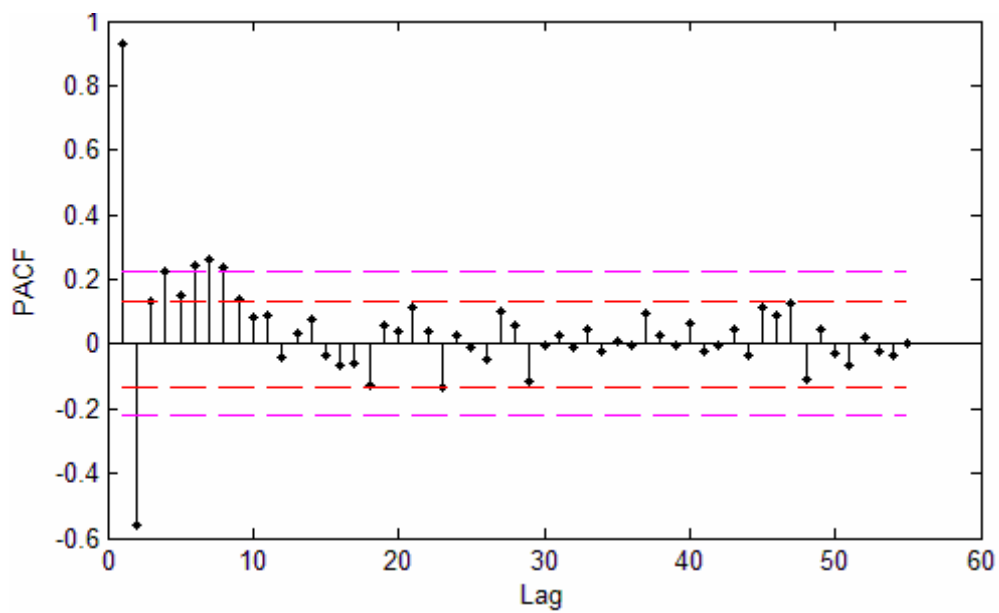

**Fig. 7.3.3: Sample PACF plot for sunspot series**

The sunspot series follows stationary behavior, which can be seen from the above two plots. We have also analytically verified this fact by unit root test. The models we have fitted to the sunspot series are AR(9), ANN (a $4\times 4\times 1$ structure) and SVM.

The obtained performance measures for the sunspot series are shown in the Table 7.3:

**Table 7.3: Forecast results for sunspot series**

| Method | MSE | MAD | RMSE | MAPE | Theil's U Statistics |
|---|---|---|---|---|---|
| AR(9) | 483.561260 | 17.628101 | 21.990026 | 60.042080% | 0.003703 |
| ANN | 334.173011 | 13.116898 | 18.280400 | 30.498342% | 0.003315 |
| SVM $\begin{pmatrix} \sigma = 290.1945, \\ C = 43.6432, \\ n = 9, N = 212 \end{pmatrix}$ | 792.961254 | 18.261674 | 28.159568 | 40.433136% | 0.004236 |

From the above table we can see that the forecasting performance of ANN is best in our experiments for the sunspot series. We have transformed the series in the range [1 2] while fitting ANN model and in the range [100 400] while fitting SVM model to achieve good forecast results. The optimal SVM hyper-parameters obtained by crossvalidation for the transformed sunspot series are shown in Table 7.3. The three forecast diagrams for the sunspot series are shown in the figures below:

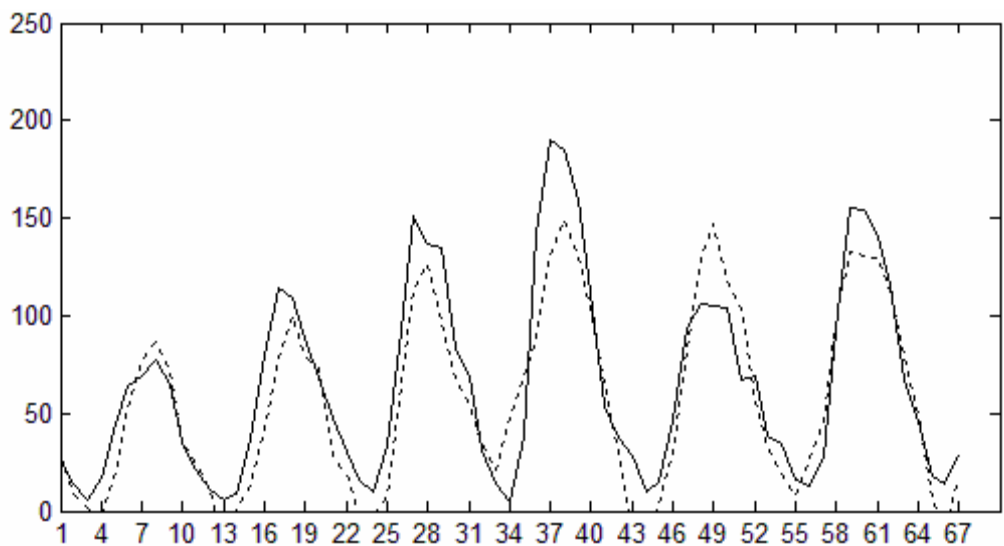

**(a) AR(9) forecast diagram**

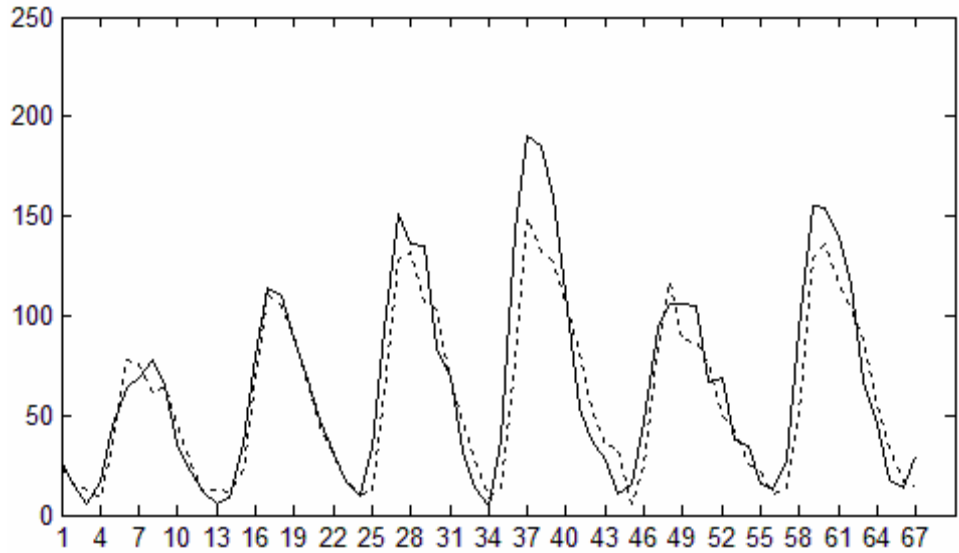

**(b) ANN forecast diagram**

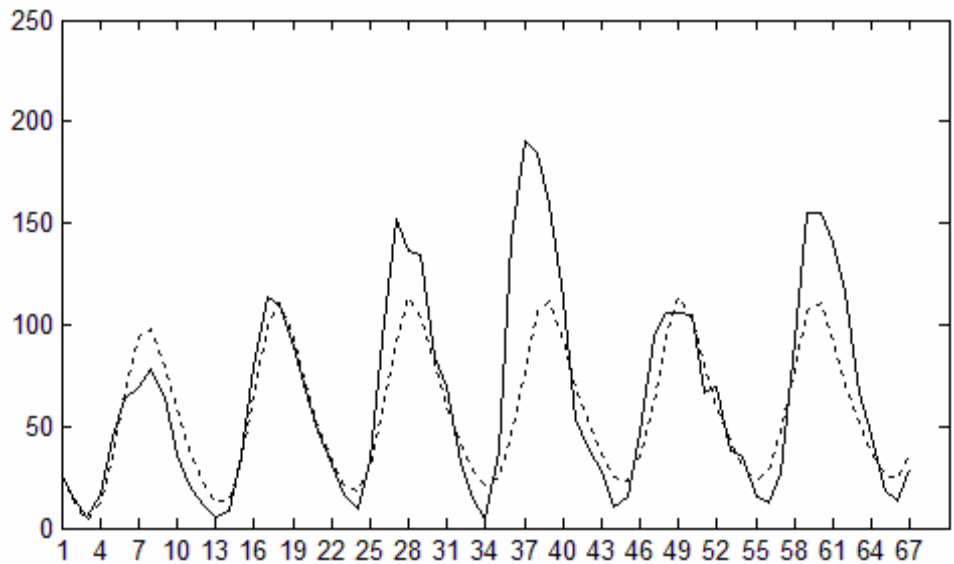

**(c) SVM forecast diagram**

**Fig. 7.3.4: Forecast diagrams for sunspot series**

From the three forecast diagrams presented in Fig. 7.3.4, we can have a graphical comparison between the actual and forecasted observations for the sunspot series, obtained by using three different techniques. It can be seen that in Fig. 7.3.4 (b) the ANN forecasted series closely resembles with the original one. However in case of SVM prediction a significant deviation between the forecasted and test values can be seen from Fig. 7.3.4 (c).

**7.4 The Airline Passenger Dataset**

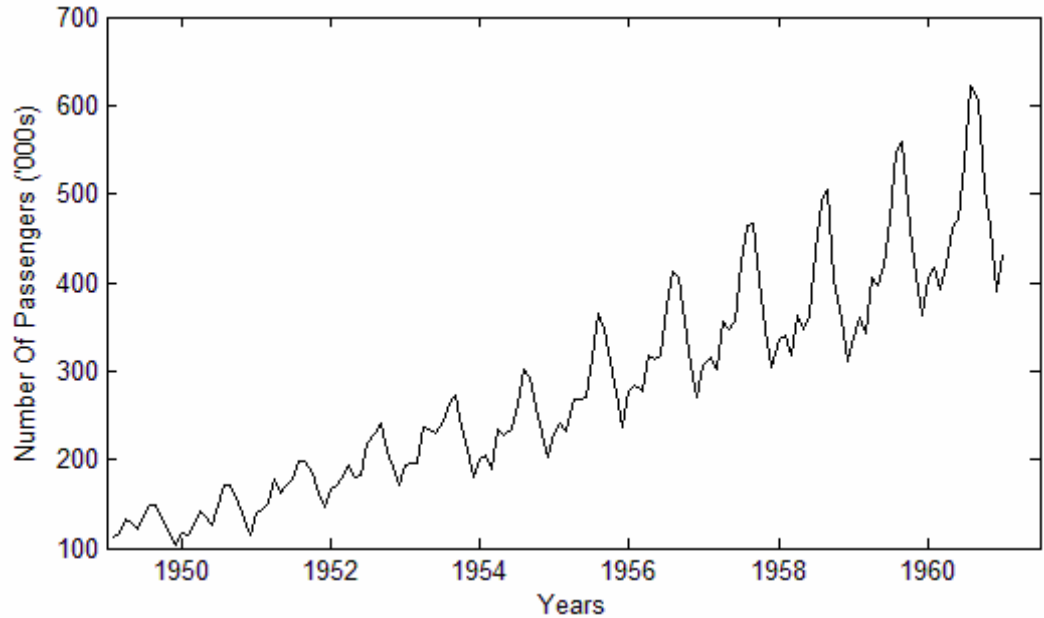

**Fig. 7.4.1: Airline passenger data series (Jan. 1949-Dec. 1960)**

The airline passenger series, shown in Fig. 7.4.1 represents the monthly total number of international airline passengers (in thousands) from January 1949 to December 1960 [3, 11, 13]. It is a famous time series and is used by many analysts including Box and Jenkins [6]. The important characteristic of this series is that it follows a multiplicative seasonal pattern with an upward trend, as can be seen from Fig. 7.4.1. The airline passenger series has total 144 observations, out of which we have used the first 132 for training and the remaining 12 for testing. Due to the presence of strong seasonal variations, the airline data is non-stationary. This can also be clarified from its sample ACF and PACF plots, as given below:

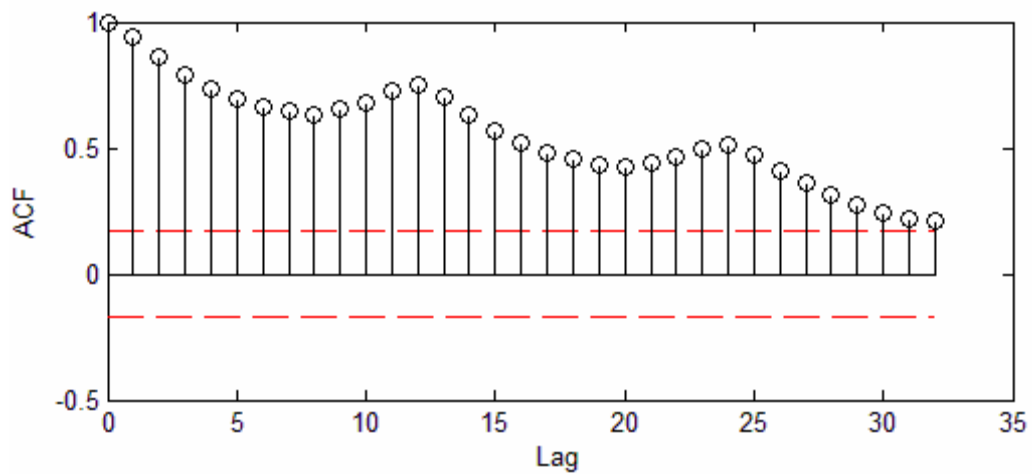

**Fig. 7.4.2: Sample ACF plot for airline series**

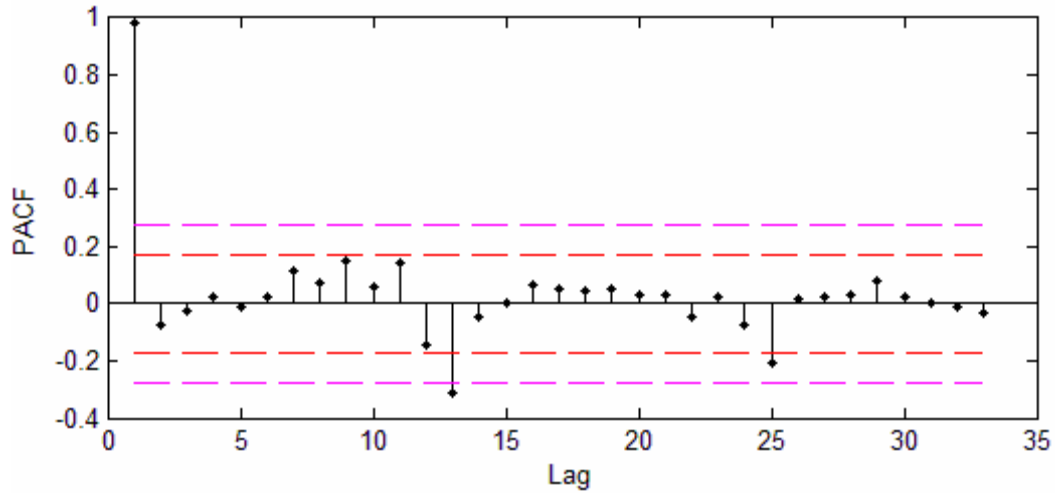

**Fig. 7.4.3: Sample PACF plot for airline series**

Fig. 7.4.2 shows that there are significantly large sample ACF values at the increasing lags, which do not diminish quickly. This indicates the non-stationarity of the airline data [6, 23]. Box et al. [6] used natural logarithmic transformation and seasonal differencing to remove non-stationarity from the series. Then they have fitted the $SARIMA(0,1,1) \times (0,1,1)^{12}$ model, which according to them is the best stochastic model for the airline data. Due to this reason $SARIMA(0,1,1) \times (0,1,1)^{12}$ is often termed as the *Airline Model* [11]. Faraway and Chatfield [11] used thirteen different ANN structures to the airline data and compared the results obtained. Also C. Hamzacebi in 2008 [3] experimented by fitting three SANN models to this data.

In this work we have fitted three different ANN, three SANN, $SARIMA(0,1,1) \times (0,1,1)^{12}$ and SVM models to the airline data. Following Box et al. [6] we have also used natural logarithmic transformation and seasonal differencing to the data for fitting the SARIMA model. Also following C. Hamzacebi [3] and Faraway et al. [11] we have rescaled the data after dividing by 100 for fitting ANN and SANN models. However for fitting the SVM model we have used the original airline data.

The forecast performance measures, we obtained for the airline data by using the above mentioned models are presented in Table 7.4. As usual the optimal SVM hyper-parameters are shown in bracket in this table. These measures are calculated using the untransformed test and forecasted observations.

## Table 7.4: Forecast results for airline passenger time series

| Method | MSE | MAD | RMSE | MAPE | Theil's U Statistics |
|---|---|---|---|---|---|
| SARIMA$(0,1,1) \times (0,1,1)^{12}$ | 189.333893 | 10.539463 | 13.759865 | 2.244234% | 0.000060 |
| ANN (1, 12; 2) | 285.633562 | 15.225263 | 16.900697 | 3.234460% | 0.000074 |
| ANN (1, 2, 12; 2) | 248.794863 | 14.159554 | 15.773232 | 3.025739% | 0.000068 |
| ANN (1, 12 13; 2) | 2532.238561 | 41.438166 | 50.321353 | 8.454268% | 0.000232 |
| SANN (1 hidden node) | 676.481142 | 24.311987 | 26.009251 | 5.138674% | 0.000118 |
| SANN (2 hidden nodes) | 275.720525 | 11.888831 | 16.604834 | 2.486088% | 0.000071 |
| SANN (3 hidden nodes) | 556.054822 | 17.693719 | 23.580815 | 3.624587% | 0.000098 |
| SVM $\begin{pmatrix} \sigma = 1.5195 \times 10^7, \\ C = 1.2767 \times 10^{10}, \\ n = 35, N = 97 \end{pmatrix}$ | 176.885301 | 10.849932 | 13.299823 | 2.336608% | 0.000057 |

From Table 7.4 we can see that SARIMA and SVM generated quite good forecasts for the airline data. Except MAD, all other performance measures obtained by SVM are least relative to the other applied models. ANN (1, 12; 2), ANN (1, 2, 12; 2) and SANN with 2 hidden nodes also produced reasonably well forecasts.

Below we present the four forecast diagrams obtained using SARIMA$(0,1,1) \times (0,1,1)^{12}$, ANN (1, 2, 12; 2), SANN (2 hidden nodes) and SVM for the airline data:

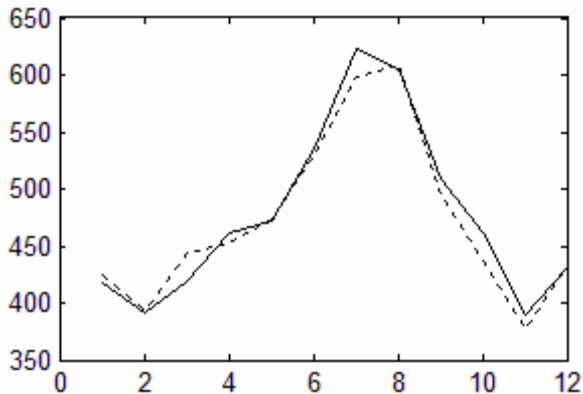
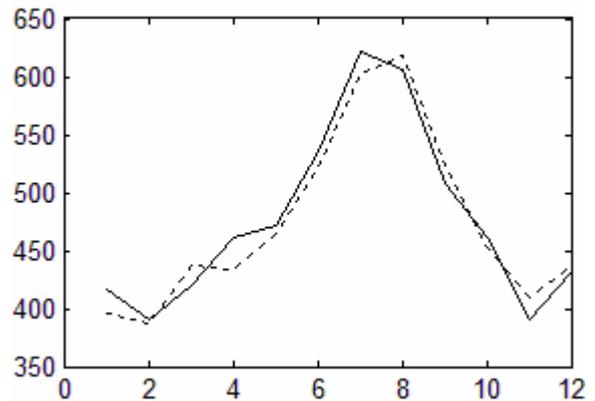

**(a) SARIMA forecast diagram**  **(b) ANN (1, 2, 12; 2) forecast diagram**

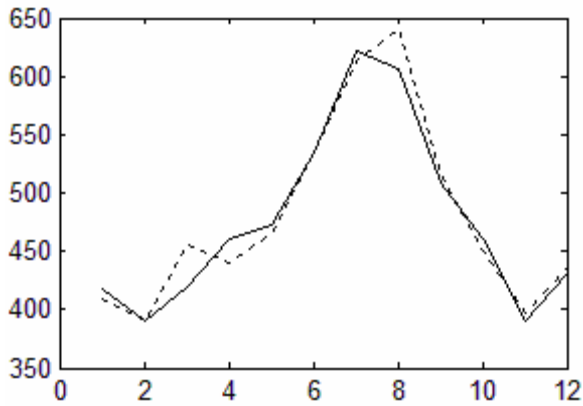
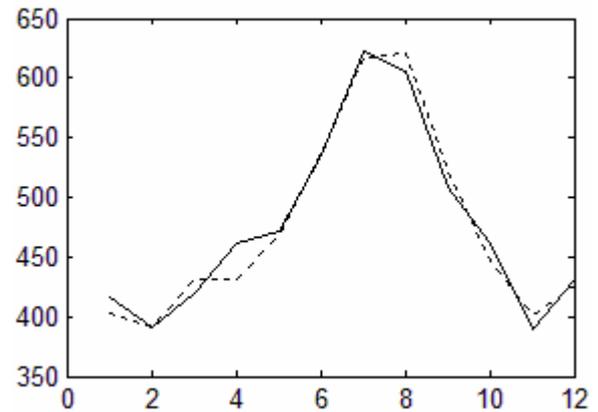

(c) SANN forecast diagram             (d) SVM forecast diagram

**Fig. 7.4.4: Forecast diagrams for airline passenger series**

The four forecast diagrams in Fig. 7.4.4 shows the success of our applied techniques to produce forecasts for the airline data. In particular Fig. 7.4.4 (d) depicts the excellent forecasting performance of SVM for this dataset.

## 7.5 The Quarterly Sales Dataset

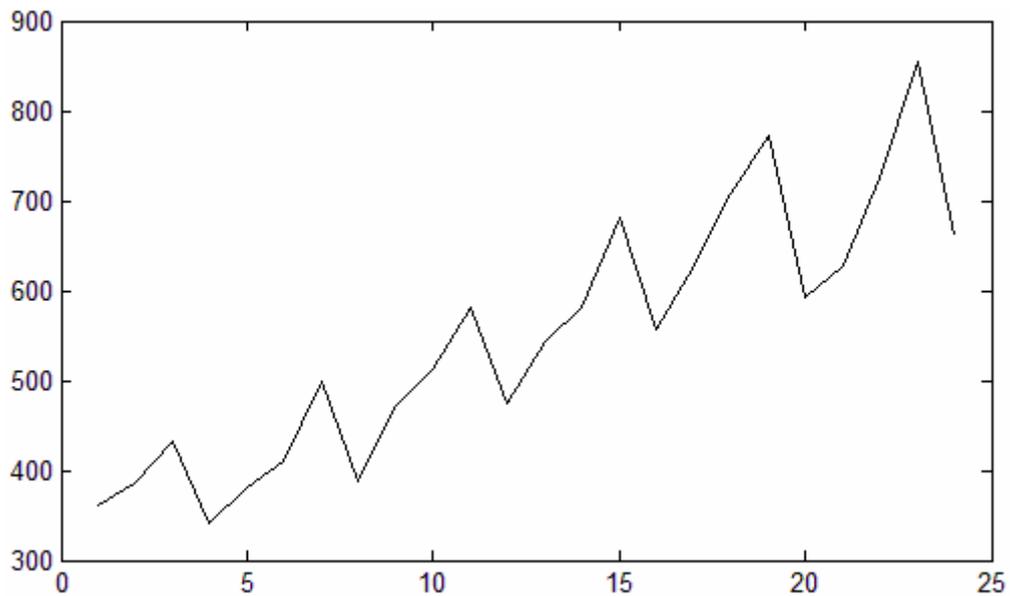

**Fig. 7.5.1: Quarterly sales time series (for 6 years)**

Fig. 7.5.1 represents the quarterly export data of a French firm for six years [3]. It is a non-stationary seasonal time series (with seasonal period 4) and was used by C. Hamzacebi [3] to assess the performance of SANN for quarterly data. This dataset has total 24 observations.

Following C. Hamzacebi [3] we have kept the first five years observations from the quarterly sales dataset for training and the sixth year observations for testing. We have fitted the $SARIMA(0,1,1) \times (0,1,1)^4$ model to the quarterly sales series after making natural logarithmic transformation and seasonal differencing. We have also fitted three SANN models (varying the number of hidden nodes to 1, 2 and 3) and the SVM model to this data. For fitting the SANN models we have used the rescaled quarterly sales data, obtained after diving by 100; for fitting the SVM model we have used the original data.

Our obtained forecast performance measures, calculated on original scale for the quarterly sales time series are presented in Table 7.5:

**Table 7.5: Forecast results for quarterly sales time series**

| Method | MSE | MAD | RMSE | MAPE | Theil's U Statistics |
|---|---|---|---|---|---|
| $SARIMA(0,1,1) \times (0,1,1)^4$ | 116.126604 | 9.740500 | 10.776205 | 1.335287% | 0.000021 |
| SANN (1 hidden node) | 968.677397 | 29.368519 | 31.123583 | 4.172137% | 0.000060 |
| SANN (2 hidden nodes) | 1004.001275 | 29.481937 | 31.685979 | 4.211440% | 0.000060 |
| SANN (3 hidden nodes) | 466.920955 | 16.734615 | 21.608354 | 2.525232% | 0.000041 |
| SVM $\begin{pmatrix} \sigma = 1.091 \times 10^4, \\ C = 7.5833 \times 10^{10}, \\ n = 10, N = 10 \end{pmatrix}$ | 1645.392593 | 32.921002 | 40.563439 | 4.902750% | 0.000075 |

From Table 7.5 we it can be seen that the best forecasts for the quarterly sales series are obtained by fitting $SARIMA(0,1,1) \times (0,1,1)^4$ model. The forecasting performances of SANN models are moderate, while those of the SVM model are not up to the expectation. We can suggest that the forecasting performance of SVM may be improved by choosing some appropriate data transformation or rescaling for this time series.

In Fig. 7.5.2 we present the SARIMA and SANN (3 hidden nodes) forecast diagrams for the quarterly sales time series:

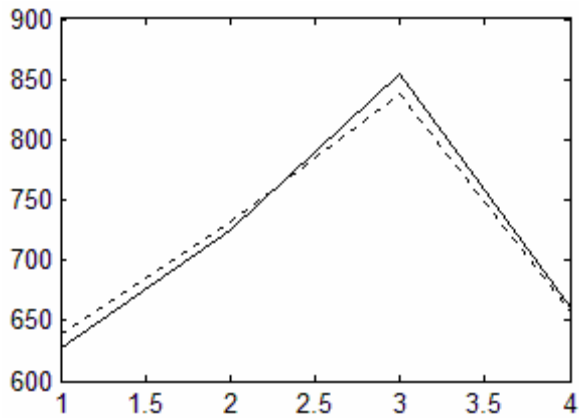 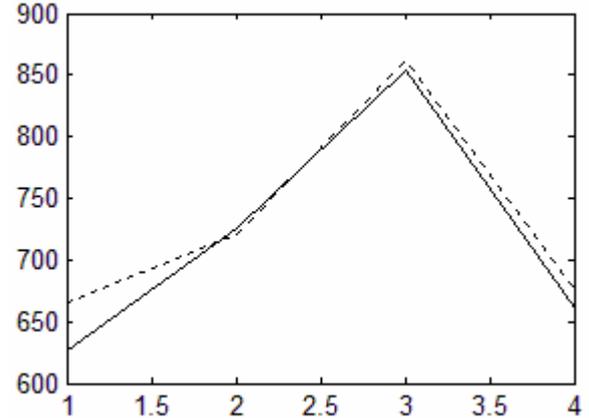

(a) SARIMA forecast diagram          (b) SANN forecast diagram

Fig. 7.5.2: Forecast diagrams for quarterly sales series

From the above two forecast diagrams we can get a visual idea about the forecasting accuracy of the mentioned SARIMA and SANN models for the quarterly sales series.

**7.6 The Quarterly U.S. Beer Production Dataset**

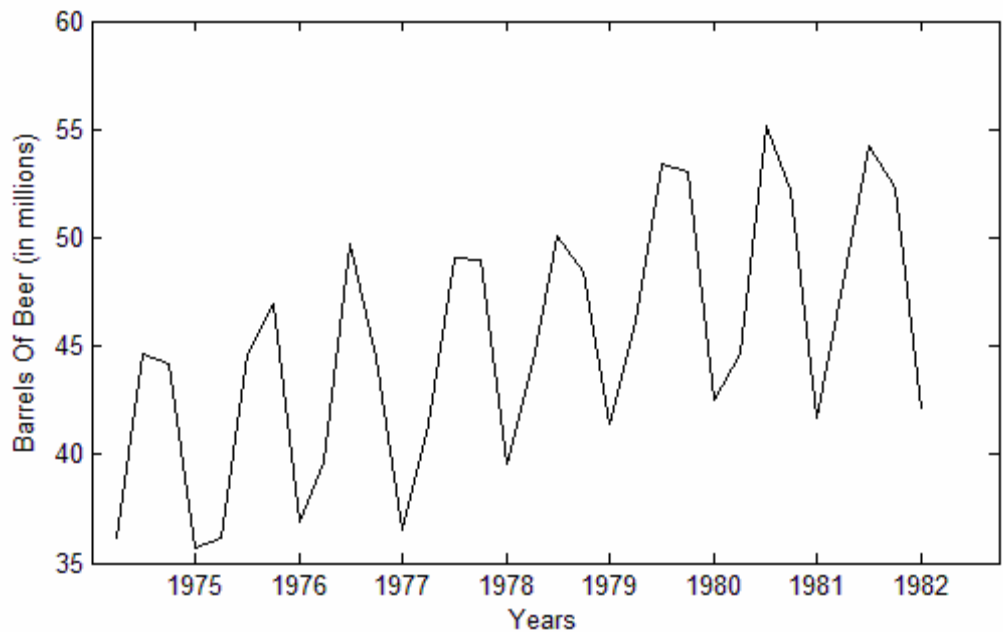

Fig. 7.6.1: Quarterly U.S. beer production time series (1975-1982)

Fig. 7.6.1 shows the quarterly U.S. beer production in millions of barrels from 1975 to 1982. This time series dataset is provided by William W. S. Wei; the associated website link is provided at the end of this ongoing book. Out of the total 32 observations in this time series, we have used the first 24 (i.e. 1975 to 1980) for training and the remaining 8 (i.e. 1981 to 1982) for testing. The forecasting models, we have fitted to this time series are $SARIMA(0,1,1) \times (0,1,1)^4$, SANN (with hidden nodes 1, 2, 3 and 4) and SVM models. As usual for fitting the SARIMA model, we have performed natural logarithmic transformation and seasonal differencing of the data values. Also for fitting SANN models, we have divided the original observations by 10.

Our obtained forecast performance measures for the quarterly U.S. beer production time series are presented in original scale in Table 7.6:

**Table 7.6: Forecast results for quarterly U.S. beer production time series**

| Method | MSE | MAD | RMSE | MAPE | Theil's U Statistics |
|---|---|---|---|---|---|
| $SARIMA(0,1,1) \times (0,1,1)^4$ | 1.784946 | 1.195592 | 1.336019 | 2.468494% | 0.000553 |
| SANN (1 hidden node) | 2.448568 | 1.307804 | 1.564790 | 2.630288% | 0.000660 |
| SANN (2 hidden nodes) | 1.868804 | 1.144692 | 1.367042 | 2.364879% | 0.000572 |
| SANN (3 hidden nodes) | 1.810862 | 1.076623 | 1.345683 | 2.325204% | 0.000556 |
| SANN (4 hidden nodes) | 1.409691 | 1.015860 | 1.187304 | 2.072790% | 0.000496 |
| SVM $\begin{pmatrix} \sigma = 33.1633, \\ C = 50.3518, \\ n = 19, N = 5 \end{pmatrix}$ | 1.511534 | 1.020438 | 1.229445 | 2.187802% | 0.000510 |

Table 7.6 shows that the relatively best forecast performance measures for quarterly U.S. beer production time series are obtained using SANN model with four hidden nodes. The performances of SARIMA and SVM models are also quite good, as can be seen from the table.

We present the three forecast diagrams for quarterly U.S. beer production time series, corresponding to SARIMA, SANN (4 hidden nodes) and SVM models in Fig. 7.6.2:

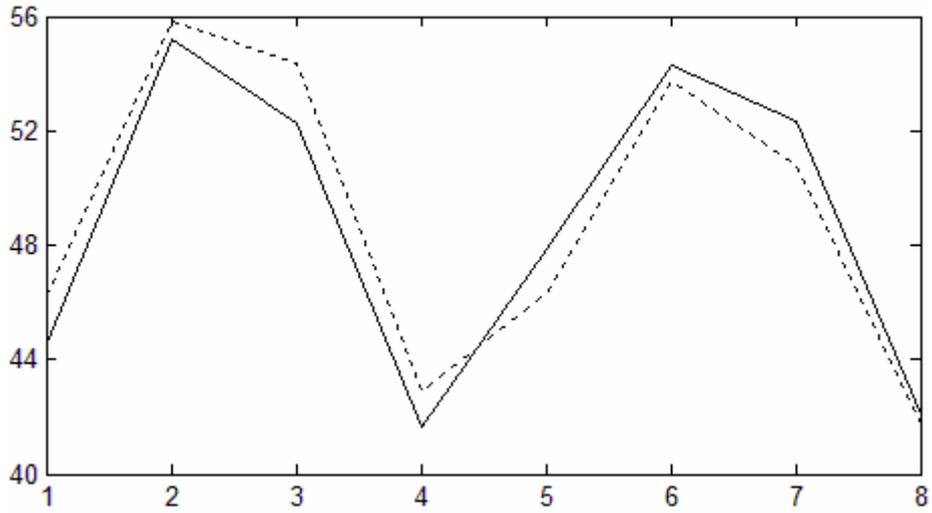

(a) SARIMA forecast diagram

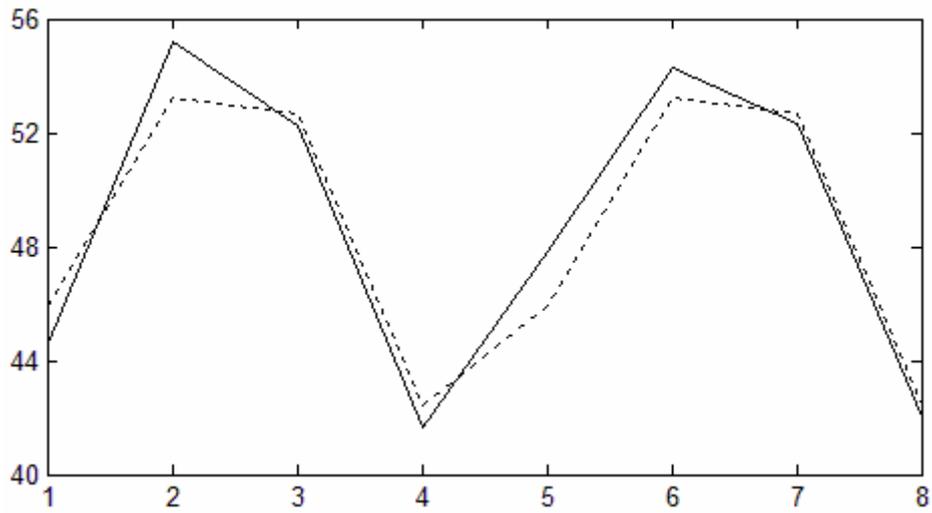

(b) SANN forecast diagram

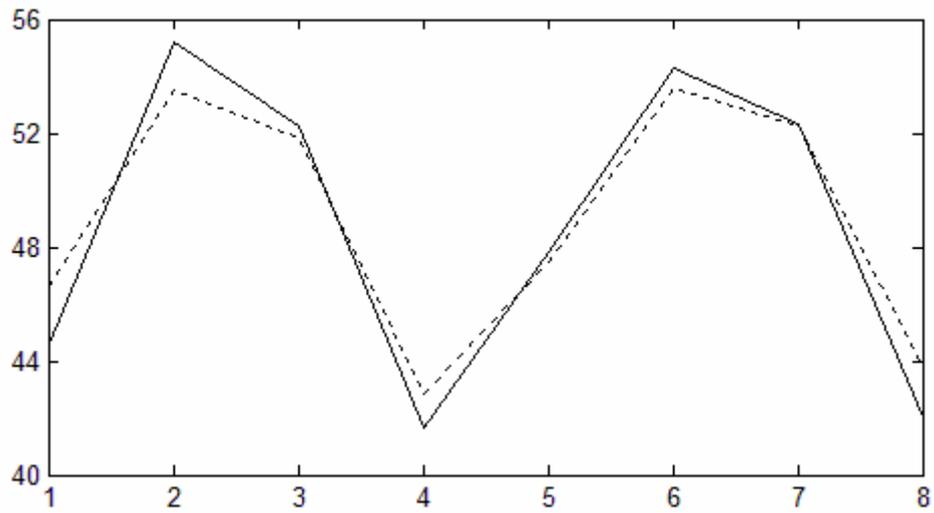

(c) SVM forecast diagram

**Fig. 7.6.2: Forecast diagrams for quarterly U.S. beer production series**

**7.7 The Monthly USA Accidental Deaths Dataset**

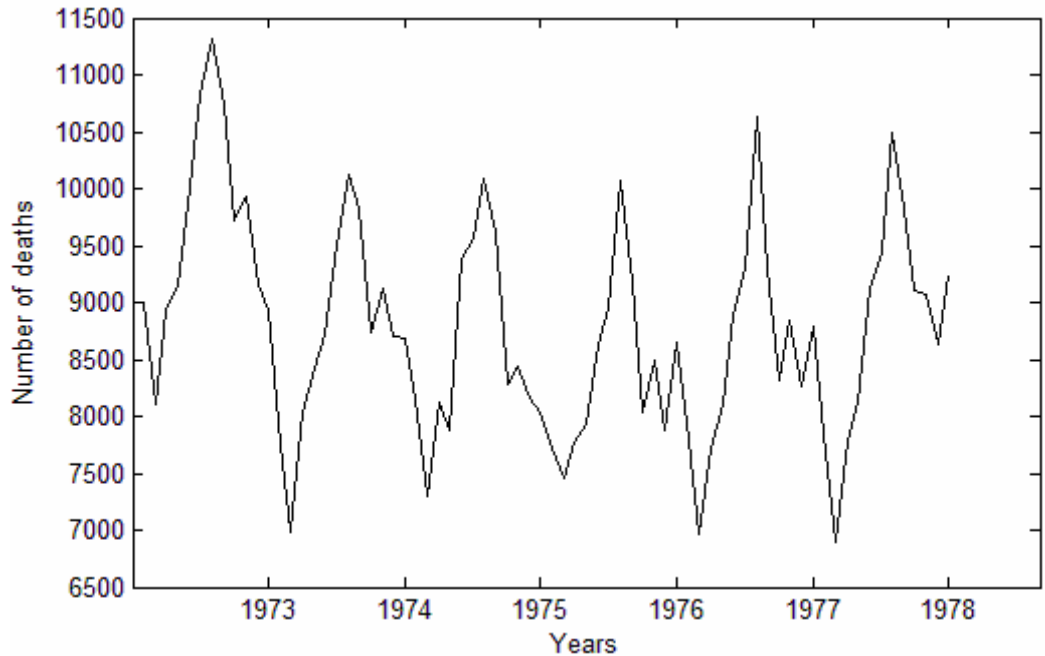

**Fig. 7.7.1: Monthly USA accidental deaths time series (1973-1978)**

The time series plot in Fig 7.7.1 represents the number of accidental deaths in USA from 1973 to 1978 on a monthly basis. This series is made available by Brockwell and Davis. We have given the associated website source at the end of this book. From the depicted Fig 7.7.1, it can be seen that the series shows seasonal fluctuations with somewhat a constant trend.

Out of the total 72 observations in the series, we have used the observations of first five years for training and those of the sixth year for testing. Thus the first 60 observations are considered for training and the remaining 12 for testing. Our fitted models for this time series are SARIMA$(0,1,1) \times (0,1,1)^{12}$, SANN (with 4 hidden nodes) and SVM. As before for fitting SARIMA model we have applied logarithmic transformation and seasonal differencing. Also For fitting SANN and SVM models the observations are divided by 100.

We present the obtained forecast performance measures in Table 7.7. These measures are calculated based on the rescaled dataset (i.e. after dividing by 100).

Table 7.7: Forecast results for monthly USA accidental deaths time series

| Method | MSE | MAD | RMSE | MAPE | Theil's U Statistics |
|---|---|---|---|---|---|
| SARIMA$(0,1,1)\times(0,1,1)^{12}$ | 3.981148 | 1.499827 | 1.995281 | 1.694144% | 0.000254 |
| SANN (4 hidden nodes) | 13.994408 | 2.986875 | 3.740910 | 3.344423% | 0.000491 |
| SVM $\begin{pmatrix} \sigma = 45.8047, \\ C = 6.8738, \\ n = 13, N = 47 \end{pmatrix}$ | 16.203377 | 3.328154 | 4.025342 | 3.708167% | 0.000531 |

From Table 7.7 we see that the minimum performance measures are obtained by the fitted SARIMA$(0,1,1)\times(0,1,1)^{12}$ model. Here to fit the SANN model 4 hidden nodes are considered, because our experiments have shown that with other number of hidden nodes the performance measures increased for this dataset. It can be suggested that the SVM performance for this dataset may be improved by performing some other suitable data preprocessing or using some other kernel function.

The two forecast diagrams for monthly USA accidental deaths time series, corresponding to the fitted SARIMA and SANN models are presented in Fig. 7.7.2:

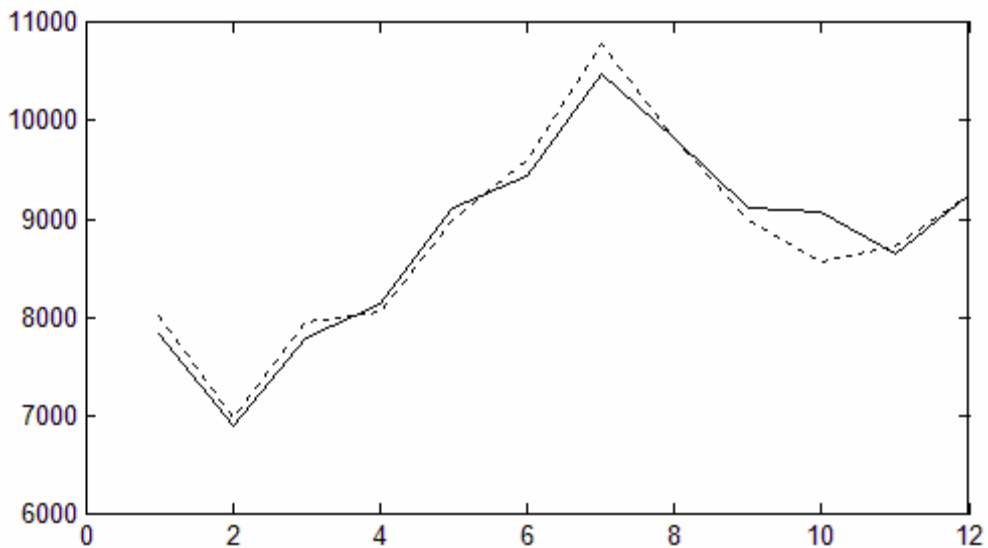

(a) SARIMA forecast diagram

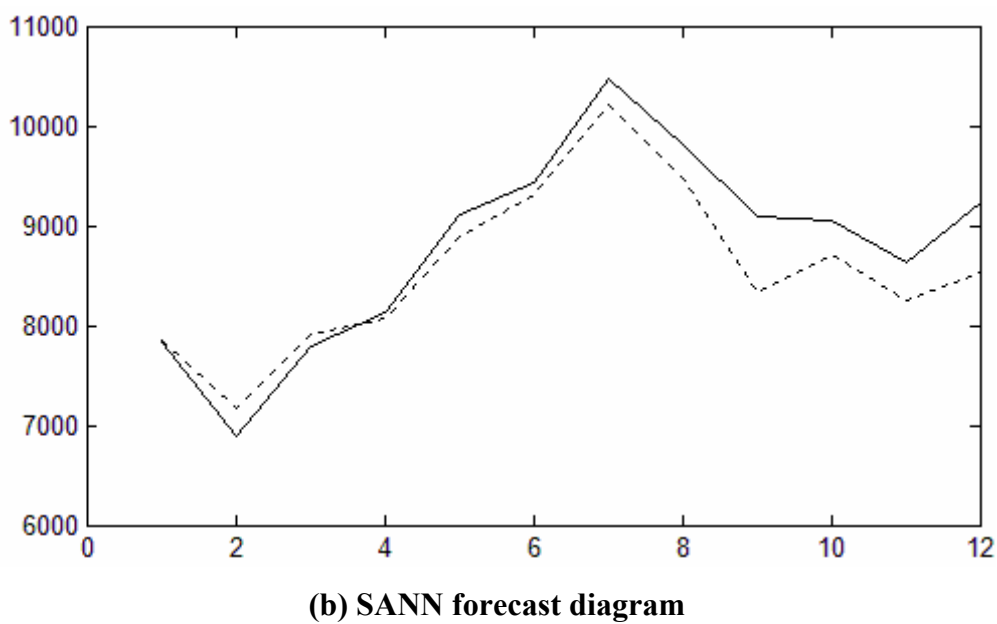

**(b) SANN forecast diagram**

**Fig. 7.7.2: Forecast diagrams for monthly USA accidental deaths series**

The excellent performance of SARIMA for the monthly USA accidental deaths time series can be visualized from the forecast diagram in Fig. 7.7.2 (a).

We have presented the forecasting results of all the experiments done by us. From the performance measures obtained for each dataset, one can have a relative idea about the effectiveness and accuracy of the fitted models. The six time series datasets, we have considered are taken from non-confidential sources and each of them is freely available for analysis. At the end of the current book, we have listed the original online source for each dataset, together with the corresponding website link from which it is collected.

# Conclusion

Broadly speaking, in this introductory book we have presented a state-of-the-art of the following popular time series forecasting models with their salient features:

- The Box-Jenkins or ARIMA models for linear time series forecasting.
- Some non-linear stochastic models, such as NMA, ARCH.
- Neural network forecasting models; TLNN and SANN.
- SVM based forecasting models; LS-SVM and DLS-SVM.

It has been seen that, the proper selection of the model orders (in case of ARIMA), the number of input, hidden and output neurons (in case of ANN) and the constant hyper-parameters (in case of SVM) is extremely crucial for successful forecasting. We have discussed the two important functions, viz. AIC and BIC, which are frequently used for ARIMA model selection. For selecting the number of appropriate neurons in ANN and constant hyper-parameters in SVM, crossvalidation should be carried out, as mentioned earlier.

We have considered a few important performance measures for evaluating the accuracy of forecasting models. It has been understood that for obtaining a reasonable knowledge about the overall forecasting error, more than one measure should be used in practice. The last chapter contains the forecasting results of our experiments, performed on six real time series datasets. Our satisfactory understanding about the considered forecasting models and their successful implementation can be observed form the five performance measures and the forecast diagrams, we obtained for each of the six datasets. However in some cases, significant deviation can be seen among the original observations and our forecasted values. In such cases, we can suggest that a suitable data preprocessing, other than those we have used in our work may improve the forecast performances.

Time series forecasting is a fast growing area of research and as such provides many scope for future works. One of them is the *Combining Approach*, i.e. to combine a number of different and dissimilar methods to improve forecast accuracy. A lot of works have been done towards this direction and various combining methods have been proposed in literature [8, 14, 15, 16]. Together with other analysis in time series forecasting, we have thought to find an efficient combining model, in future if possible. With the aim of further studies in time series modeling and forecasting, here we conclude the present book.

# References

[1]    Burges, C.J.C., "A tutorial on support vector machines for pattern recognition", Data Mining and Knowledge Discovery, 2 (1998), pages: 121-167.

[2]    C. Chatfield, "Model uncertainty and forecast accuracy", J. Forecasting 15 (1996), pages: 495–508.

[3]    C. Hamzacebi, "Improving artificial neural networks' performance in seasonal time series forecasting", Information Sciences 178 (2008), pages: 4550-4559.

[4]    F. Girosi, M. Jones, and T. Poggio, "Priors, stabilizers and basis functions: From regularization to radial, tensor and additive splines." AI Memo No: 1430, MIT AI Lab, 1993.

[5]    G. Zhang, B.E. Patuwo, M.Y. Hu, "Forecasting with artificial neural networks: The state of the art", International Journal of Forecasting 14 (1998), pages: 35-62.

[6]    G.E.P. Box, G. Jenkins, "Time Series Analysis, Forecasting and Control", Holden-Day, San Francisco, CA, 1970.

[7]    G.P. Zhang, "A neural network ensemble method with jittered training data for time series forecasting", Information Sciences 177 (2007), pages: 5329–5346.

[8]    G.P. Zhang, "Time series forecasting using a hybrid ARIMA and neural network model", Neurocomputing 50 (2003), pages: 159–175.

[9]    H. Park, "Forecasting Three-Month Treasury Bills Using ARIMA and GARCH Models", Econ 930, Department of Economics, Kansas State University, 1999.

[10]   H. Tong, "Threshold Models in Non-Linear Time Series Analysis", Springer-Verlag, New York, 1983.

[11]   J. Faraway, C. Chatfield, "Time series forecasting with neural networks: a comparative study using the airline data", Applied Statistics 47 (1998), pages: 231–250.

[12]   J. Lee, "Univariate time series modeling and forecasting (Box-Jenkins Method)", Econ 413, lecture 4.

[13]   J.M. Kihoro, R.O. Otieno, C. Wafula, "Seasonal Time Series Forecasting: A Comparative Study of ARIMA and ANN Models", African Journal of Science and Technology (AJST) Science and Engineering Series Vol. 5, No. 2, pages: 41-49.

[14]   J. Scott Armstrong, "Combining Forecasts: The End of the Beginning or the Beginning of the End?", International Journal of Forecasting 5 (1989), pages: 585-588.

[15]  J. Scott Armstrong, "Combining Forecasts", Principles of Forecasting: A Handbook for Researchers and Practitioners; J. Scott Armstrong (ed.): Norwell, MA: Kluwer Academic Publishers, 2001.

[16]  J. Scott Armstrong, "Findings from evidence-based forecasting: Methods for reducing forecast error", International Journal of Forecasting 22 (2006), pages: 583–598.

[17]  J.W. Galbraith, V. Zinde-Walsh, "Autoregression-based estimators for ARFIMA models", CIRANO Working Papers, No: 2011s-11, Feb 2001.

[18]  J.A.K. Suykens and J. Vandewalle, "Least squares support vector machines classifiers", Neural Processing Letters, vol. 9, no. 3, pp. 293-300, June 1999.

[19]  J.A.K. Suykens and J. Vandewalle, "Recurrent least squares support vector machines". IEEE Trans. Circuits Systems-I, vol. 47, no. 7, pp. 1109 - 1114, July 2000.

[20]  Joarder Kamruzzaman, Rezaul Begg, Ruhul Sarker, "Artificial Neural Networks in Finance and Manufacturing", Idea Group Publishing, USA.

[21]  John H. Cochrane, "Time Series for Macroeconomics and Finance", Graduate School of Business, University of Chicago, spring 1997.

[22]  K. Hornik, M. Stinchcombe, H. White, "Multilayer feed-forward networks are universal approximators", Neural Networks 2 (1989), pages: 359–366.

[23]  K.W. Hipel, A.I. McLeod, "Time Series Modelling of Water Resources and Environmental Systems", Amsterdam, Elsevier 1994.

[24]  L.J. Cao and Francis E.H. Tay "Support Vector Machine with Adaptive Parameters in Financial Time Series Forecasting", IEEE Transaction on Neural Networks, Vol. 14, No. 6, November 2003, pages: 1506-1518.

[25]  M. Cottrell, B. Girard, Y. Girard, M. Mangeas, C. Muller, "Neural modeling for time series: a statistical stepwise method for weight elimination", IEEE Trans. Neural Networks 6 (1995), pages: 1355–1364.

[26]  M.J. Campbell, A.M. Walker, "A survey of statistical work on the MacKenzie River series of annual Canadian lynx trappings for the years 1821–1934, and a new analysis", J. R. Statist. Soc. Ser. A 140 (1977), pages: 411–431.

[27]  R. Lombardo, J. Flaherty, "Modelling Private New Housing Starts In Australia", Pacific-Rim Real Estate Society Conference, University of Technology Sydney (UTS), January 24-27, 2000.

[28]  R. Parrelli, "Introduction to ARCH & GARCH models", Optional TA Handouts, Econ 472 Department of Economics, University of Illinois, 2001.

[29]  Satish Kumar, "Neural Networks, A Classroom Approach", Tata McGraw-Hill Publishing Company Limited.

[30]  T. Farooq, A. Guergachi and S. Krishnan, "Chaotic time series prediction using knowledge based Green's Kernel and least-squares support vector machines", Systems, Man and Cybernetics, 2007. ISIC. 7-10 Oct. 2007, pages: 373-378.

[31]  T. Raicharoen, C. Lursinsap, P. Sanguanbhoki, "Application of critical support vector machine to time series prediction", Circuits and Systems, 2003. ISCAS '03.Proceedings of the 2003 International Symposium on Volume 5, 25-28 May, 2003, pages: V-741-V-744.

[32]  T. Van Gestel, J.A.K. Suykens, D. Baestaens, A. Lambrechts, G. Lanckriet, B. Vandaele, B. De Moor, and J. Vandewalle, "Financial time series prediction using least squares support vector machines within the evidence framework", IEEE Trans. Neural Networks, vol. 12, no. 4, pp. 809 - 821, July 2001.

[33]  V. Vapnik, "Statistical Learning Theory", New York: Wiley, 1998.

[34]  Yugang Fan, Ping Li and Zhihuan Song, "Dynamic least square support vector machine", Proceedings of the 6[th] World Congress on Intelligent Control and Automation (WCICA), June 21-23, 2006, Dalian, China, pages: 4886-4889.

# Datasets Sources

**1. The Canadian Lynx Dataset**

   Original source: Elton, C. and Nicholson, M. (1942), "The ten year cycle in numbers of Canadian lynx", J. Animal Ecology, Vol. 11, pages: 215-244.

Website link: `http://www.stats.uwo.ca/faculty/aim/epubs/mhsets/ecology/lynx.1`

**2. The Wolf's Sunspot Dataset**

   Original source: H. Tong, "Non-linear Time Series: A Dynamical System Approach", Oxford Statistical Science, Series 6, July 1993, pages: 469-471.

Website link: `http://www.stats.uwo.ca/faculty/aim/epubs/mhsets/annual/sunspt.1`

**3. The Airline Passenger Dataset**

   Original source: R. G. Brown, "Smoothing, Forecasting and Prediction of Discrete Time Series", Prentice-Hall, Englewood Cliffs, 1994.

   This dataset is also used by Box and Jenkins in [6] and they had named it as Series G.

Website link: `http://robjhyndman.com/TSDL/data/airpass.dat`

**4. The Quarterly Sales Dataset**

   Original source: S. Makridakis, S. Wheelwright, R. J. Hyndman, "Forecasting: Methods and Applications", 3rd edition, John Wiley & Sons, New York, 1998.

Website link: `http://robjhyndman.com/forecasting/data/qsales.csv`

**5. The Quarterly U.S. Beer Production Dataset**

   Original source: William W. S. Wei, "Time Series Analysis: Univariate and Multivariate Methods", 2nd edition, Addison Wesley; July, 2005.

Website link: `http://astro.temple.edu/~wwei/datasets/W10.txt`

**6. The Monthly USA Accidental Deaths Dataset (1973-1978)**

   Original source: P. J. Brockwell, R. A. Davis, "Introduction to Time Series and Forecasting", 2nd edition, Springer Publication; March, 2003.

Website link: `http://robjhyndman.com/forecasting/data/deaths.csv`